\documentclass[runningheads]{llncs}
\usepackage{graphicx}

\usepackage{subcaption}

\usepackage{placeins}
\usepackage{amsmath,amssymb} 

\usepackage{microtype}
\usepackage{booktabs}

\usepackage{tikz}
\usepackage{comment} 
\usepackage{amsmath,amssymb} 
\usepackage{color}

\usepackage[width=122mm,left=12mm,paperwidth=146mm,height=193mm,top=12mm,paperheight=217mm]{geometry}

\begin{document}
\pagestyle{headings}
\mainmatter
\def\ECCVSubNumber{100}  

\title{Black-Box Optimization of Object Detector Scales} 

\titlerunning{Black-Box Optimization of Object Detector Scales}
%
\author{Mohandass Muthuraja\inst{1} \and
Octavio Arriaga\inst{2} \and
Paul Plöger\inst{1} \and
Frank Kirchner\inst{2} \and
Matias Valdenegro-Toro\inst{3}}
\authorrunning{M. Muthuraja et al.}
%
\institute{Hochschule Bonn-Rhein-Sieg, Sankt Augustin 53757, Germany \\
\email{mohandass.psgrobo@gmail.com} \and
University of Bremen, Bremen 28359, Germany \and
German Research Center for Artificial Intelligence, Bremen 28359, Germany\\
\email{matias.valdenegro@dfki.de}}
\maketitle

\begin{abstract}
    Object detectors have improved considerably in the last years by using advanced CNN architectures. However, many detector hyper-parameters are generally manually tuned, or they are used with values set by the detector authors. Automatic Hyper-parameter optimization has not been explored in improving CNN-based object detectors hyper-parameters. In this work, we propose the use of Black-box optimization methods to tune the prior/default box scales in Faster R-CNN and SSD, using Bayesian Optimization, SMAC, and CMA-ES.
    
    We show that by tuning the input image size and prior box anchor scale on Faster R-CNN mAP increases by 2\% on PASCAL VOC 2007, and by 3\% with SSD. On the COCO dataset with SSD there are mAP improvement in the medium and large objects, but mAP decreases by 1\% in small objects. We also perform a regression analysis to find the significant hyper-parameters to tune.
    
\end{abstract}

\section{Introduction}
Object detection deals with classifying and localizing objects of interest in a given image. In the recent years, a great deal of research has been done in object detection. Moreover, it has  multiple application domain such as autonomous cars, anomaly detection, medical image analysis, video surveillance, and so on.
Advancements in Convolutional Neural Networks (CNN) have taken deep learning based object detection a step forward as they have proved to outperform traditional computer vision methods in benchmark datasets like MS-COCO \cite{lin2014microsoft} and PASCAL VOC  \cite{Everingham10}.

The ability of CNN architectures to represent image high-level features is one of the reasons for remarkable performance in state-of-the-art object detectors \cite{zhao2019object}. However, performance depends heavily on the selection of various hyper-parameters that guide and control the learning process. Every object detection methods has several hyper-parameters like input image dimensions, size and scales of prior/default anchor boxes, multi-task loss weights, number of output proposals, in addition to conventional neural network hyper-parameters like learning rate, momentum and decay rate. The right choice of hyper-parameters is essential because it plays a significant role in the model's performance \cite{feurer_hyperparameter_2019}. 

Hyper-parameter tuning is challenging as one needs to choose the right settings from the high-dimensional search space efficiently. 
Domain experts have insight on setting hyper-parameters. Also, they conduct many experiments and choose values after many trial and error runs. Furthermore, hyper-parameters depends on the dataset as values which work fine for one dataset may not provide the same performance on a different one \cite{automl_book}. Hyper-parameter diversity is also problematic as they can be binary, categorical, continuous and conditional \cite{automl_book}.

The current growth of the machine learning field has created a need for automating this laborious process by avoiding human intervention. Automated machine learning (AutoML) \cite{automl} is a newly emerging field which aims to automate the entire machine learning process. Besides AutoML, Black-box optimization methods can also be applied for the task of hyper-parameter optimization. Most basic methods are grid search \cite{montgomery2001design} and random search \cite{bergstra2012random}. However, both methods take a considerable amount of time and are computationally expensive.

Guided search can reduce the computational complexity and the time taken to find the right set of hyper-parameters. Bayesian optimization (BO) \cite{movckus1975bayesian} is a guided way as it considers prior information. It has proved to achieve better results with fewer computations when compared to Grid  \cite{montgomery2001design} and Random search \cite{bergstra2012random} in image classification, speech recognition and natural language modeling \cite{snoek2012practical}\cite{snoek2015scalable}\cite{melis2017state}\cite{dahl2013improving}. Also, population-based approaches, namely Genetic Algorithms, Genetic Programming, Evolutionary Strategy, Evolutionary Programming and Particle Swarm Optimization have shown remarkable results \cite{coello2007evolutionary}.

All the previously discussed methods have not been applied in tuning object detection hyper-parameters. These are generally tuned manually by trial and error by the community, or used without tuning as they are defined by their authors, which might be sub-optimal for a specific dataset.

This work intends to study the applicability of Black-box optimization methods such as Bayesian Optimization \cite{movckus1975bayesian} and Co-variance Matrix Adaptation Evolution Strategy (CMA-ES)  \cite{hansen2016cma} for tuning object detection hyper-parameters, in particular the anchor/default box scales. This will allow the detector to be specifically tuned to a particular dataset, instead of being hand-tuned and producing sub-optimal performance.

To validate the proposed approach, black-box hyper-parameter optimization methods are used to optimize the Single-Shot MultiBox Detector \cite{liu2016ssd} and Faster R-CNN \cite{ren2015faster} detectors on a variety of datasets to achieve the best performance.  We find that the Black-box optimization is able to improve mAP performance and achieve better results than the hand-tuned configurations in most of the cases.

The contributions of this work are: we propose the use of black-box optimization methods to tune the prior boxes of Faster R-CNN, and the default boxes of SSD. We show that by using these methods, performance in terms of mAP on PASCAL VOC and MS-COCO increases by around 1-3\%. We also show that the scales learned with black-box optimization transfer from PASCAL VOC 2007 to VOC 2012, with an mAP improvement as well, and we perform a regression analysis to find out which are the most important hyper-parameters to tune.

It should be noted that the objective of this work is not to improve or beat state of the art detectors in many datasets, but to show the importance of automatic hyper-parameter tuning for object detection, in particular to tune the scales and input image size automatically, without a manual process. Our aim of this paper is important for real-use-cases trying to use state of the art object detectors in novel datasets, specially for non-experts in computer vision.

\section{Related Work}

\subsection{Black-Box Optimization in Deep Learning}
Black-box optimization methods are widely used in tuning hyper-parameters of deep learning algorithms. Bayesian optimization with Gaussian processes (BOGP) was first used to optimize hyper-parameters of an image recognition deep learning architecture on CIFAR-10 \cite{Krizhevsky09learningmultiple} and was able to achieve 3\% increase over the state of the art in 2012 \cite{snoek2012practical}. A new approach in Bayesian Optimization, called as Deep Networks for Global Optimization (DNGO) \cite{snoek2015scalable} uses neural networks instead of Gaussian processes (GP) as a surrogate model for fitting distributions over the objective functions. Bayesian optimization with DNGO also supports parallelism for hyper-parameter optimization. DNGO was used to tune the hyper-parameters of various deep learning problems such as image classification and image caption generation.

In the work presented in \cite{loshchilov2016cma}, CMA-ES is compared with Bayesian optimization for hyper-parameter optimization of deep convolutional neural networks for MNIST classification problem. The Particle Swarm Optimization (PSO) algorithm has also proved beneficial for a hyper-parameter optimization problem. In the work presented in \cite{lorenzo2017particle} and \cite{lorenzo2017hyper}, PSO is used for optimizing the hyper-parameters of the Deep Neural Network (DNN) in parallel and quickly on the CIFAR-10 dataset. \cite{feurer_hyperparameter_2019} has discussed various notable strategies in hyper-parameter optimization. However, these methods have not been explored much in tuning the hyper-parameters specific to deep learning-based object detectors. In this work, we focus on the area of automatic tuning of object detector hyper-parameters using black-box optimization methods.

\subsection{Object Detection Hyper-parameters Selection}
The designing of prior/default anchor boxes is a big challenge in object detection. The design process completely depends on the size and ratio of objects to be detected in a particular dataset. \cite{zhong2018anchor} proposed an approach to dynamically adapt the design of anchor boxes using the gradients of the loss function. The anchor box optimization method was integrated into YOLO v2 \cite{redmon2017yolo9000} and had obtained a 1\% mAP gain in MS-COCO \cite{lin2014microsoft} and PASCAL VOC \cite{Everingham10} datasets.

In YOLOv2 \cite{redmon2017yolo9000}, k-Means clustering is used to determine the prior anchor boxes. The ground truth bounding boxes in the training set are clustered based on the Intersection Over Union (IOU) scores instead of the conventional Euclidean distance, as it produces error for larger boxes than smaller ones. Finally, the cluster centroids are used for designing the anchor boxes. Our approach has an advantage over these methods as it is not constrained only to anchor boxes and can include other object detection hyper-parameters. 

\section{Proposed Approach}
In our experiments we use Bayesian optimization and CMA-ES for tuning the object detection hyperparameters. In this section we will briefly discuss about Bayesian Optimization \cite{movckus1975bayesian} and CMA-ES \cite{hansen2016cma}.
\subsection{Bayesian Optimization}
Bayesian optimization (BO) \cite{movckus1975bayesian} is an efficient and effective method for global optimization problems. Bayesian optimization over the years has also evolved as a prominent solution for hyperparameter optimization problems. In Bayesian optimization, initially a random set of hyperparameters is evaluated, and a probabilistic surrogate model is fit to this data 'D'. This probabilistic surrogate model is used by an acquisition function to compute a utility score for a different set of hyperparameters. Then the hyperparameters with a better score will be evaluated on the actual objective function, and the evaluation result is used to update the probabilistic surrogate model. The optimization process goes in iteration until a computational budget is met. 

In a nutshell, the two main components of the Bayesian optimization are the probabilistic surrogate model and acquisition function. The probabilistic surrogate model is the Bayesian approximation of the actual objective function to derive samples efficiently. The acquisition function implies a trade-off between exploration and exploitation. We chose Expected improvement (EI) \cite{mockus1978application} over other acquisition function like Probability of Improvement (PI) \cite{kushner1964new} and Gaussian process upper confidence bound (GP-UCB) \cite{kaufmann2012bayesian} as EI has proved to have the best convergence rates \cite{bull2011convergence}.

In our experiments, we use Gaussian Processes (GP) based BO, and Random Forests based BO, which is called Sequential Model-based Algorithmic Configuration (SMAC) \cite{hutter2011sequential}.

\begin{figure}[!t]
    \centering
    \begin{subfigure}[b]{0.48\textwidth}
        \centering
        \includegraphics[width=1\columnwidth]{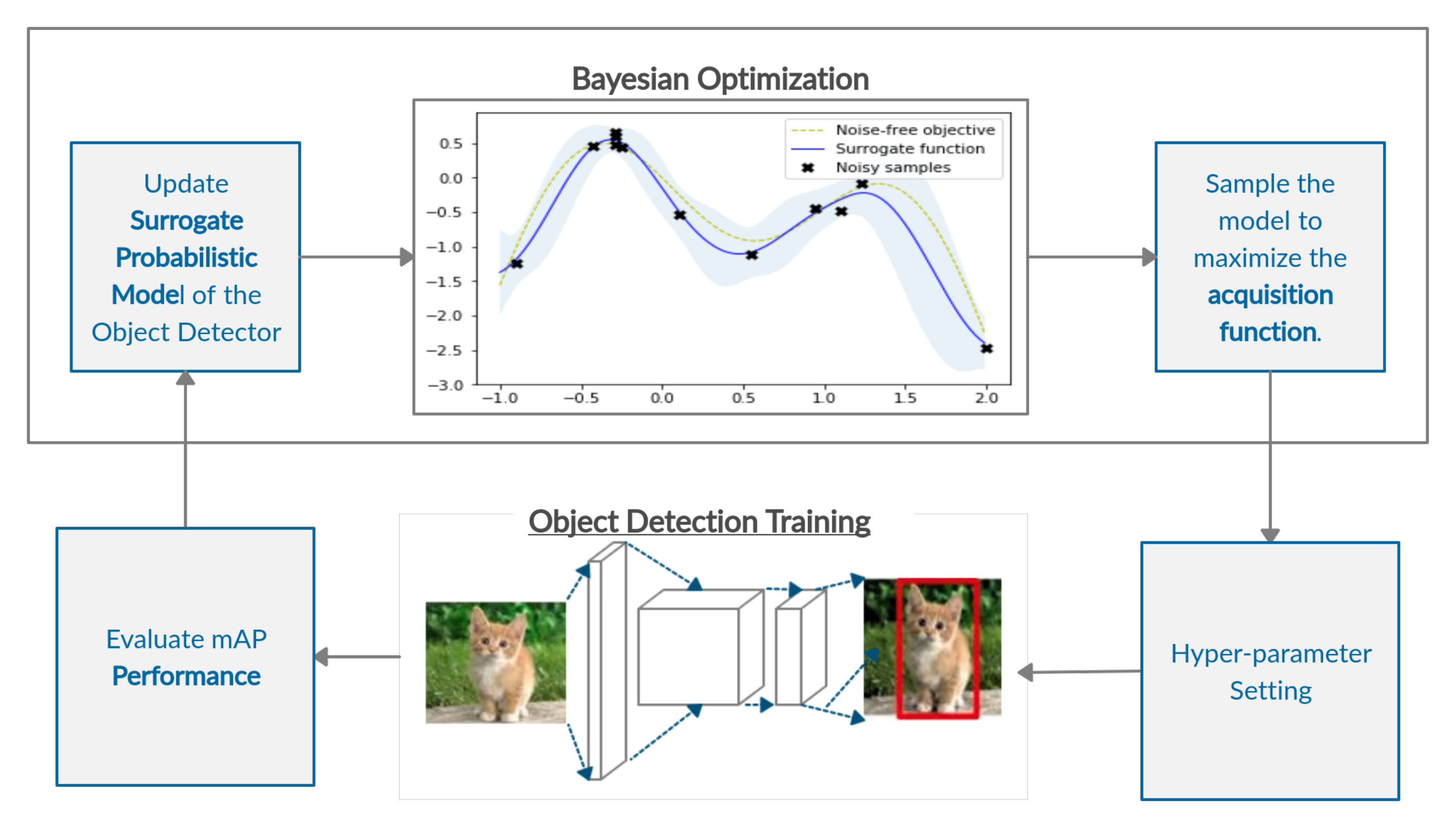}
        \caption{Bayesian Optimization}
        \label{Bayesian_Optimization_Pipeline}
    \end{subfigure} 
    ~
    \begin{subfigure}[b]{0.48\textwidth}
        \centering
        \includegraphics[width=1\columnwidth]{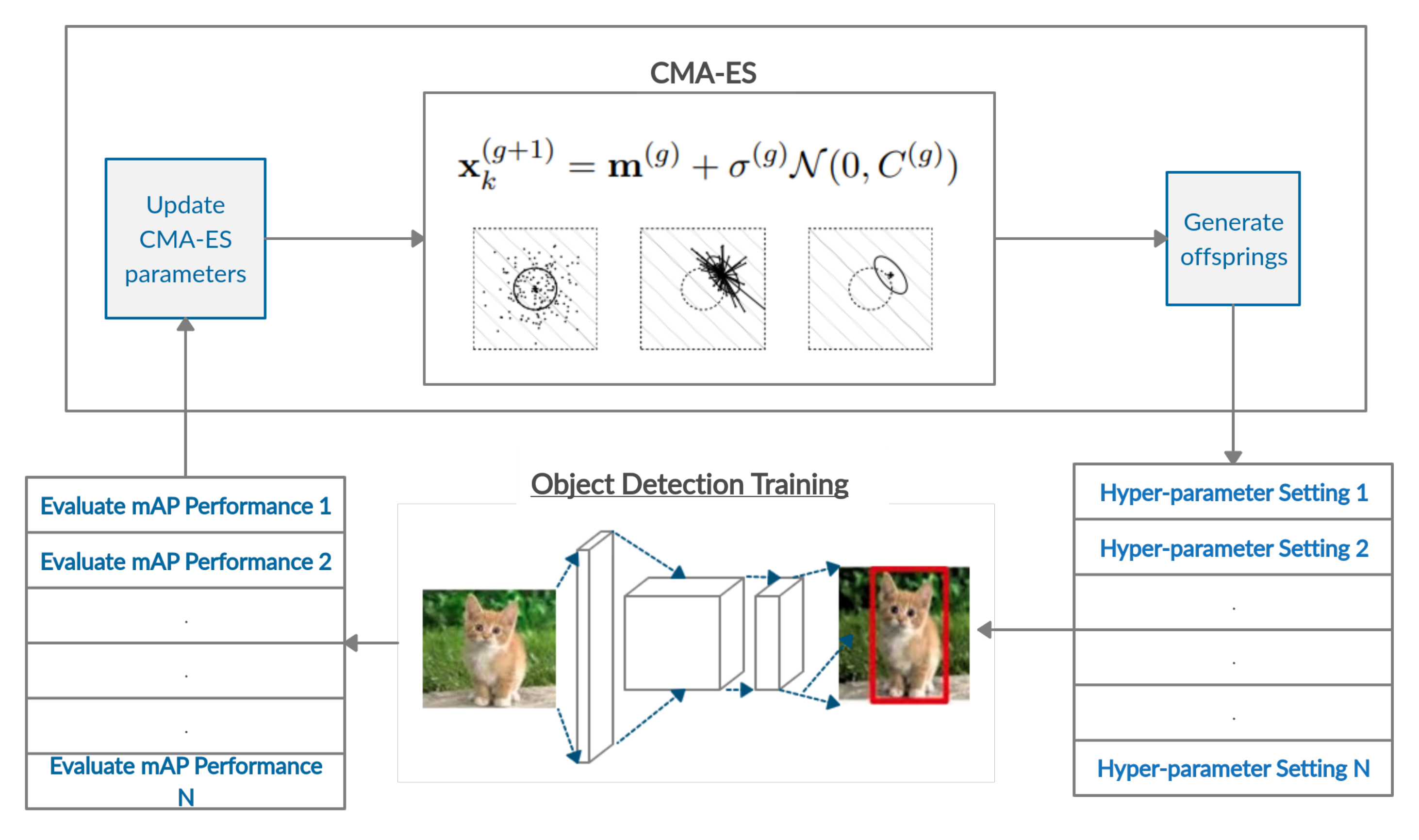}
        \caption{CMA-ES}
        \label{CMA_ES_Pipeline}
    \end{subfigure}
    \caption{Hyperparameter Optimization Pipeline for Object Detection}
    \label{CMA_ES_Pipeline}
\end{figure}

Figure \ref{Bayesian_Optimization_Pipeline} shows the flowchart of the BO pipeline for object detection used in our experiments. We use the SMAC3 \footnote{\url{https://github.com/automl/SMAC3}} \cite{smac-2017} implementation, containing BO and SMAC implementations.

\subsection{Covariance Matrix Adaptation Evolution Strategy}
Covariance Matrix Adaptation Evolution Strategy (CMA-ES) \cite{hansen2016cma} is also used for continuous hyper-parameter tuning. CMA-ES is an evolutionary algorithm used for derivative-free optimization of non-linear, non-convex continuous objective functions. The main advantage of CMA-ES over other evolutionary algorithms is that it can behave well with small population sizes.
In a nutshell, CMA-ES generate $\lambda$ samples from a multi-variate normal distribution and evaluates these samples in the objective function to obtain a fitness value for every sample. Based on the fitness of the samples, the multi-variate normal distribution is adjusted to generate new samples in the next generation (iteration). This process is repeated until a best sample is found or a computation budget is met.
Figure \ref{CMA_ES_Pipeline} shows the CMA-ES pipeline flowchart for object detection used in our experiments. We use the pycma\footnote{\url{https://github.com/CMA-ES/pycma}} CMA-ES implementation \cite{hansen2019pycma}.

\subsection{Hyper-Parameter Space Design}

In this section we describe the hyper-parameter space design for two selected object detection models.

We chose these two detectors since they perform detections using different approaches and could validate the generality of our approach. Specifically, we chose SSD and Faster R-CNN which are defined respectively as one-stage and two-stage detectors.
There are more recent detectors based on these designs, but we wanted to evaluate the original versions as to separate the effect of other improvements with hyper-parameter tuning.

\textbf{Faster R-CNN} \cite{ren2015faster} \cite{ren2016faster} uses a Region Proposal Network (RPN) for identifying regions of interest which replaces the selective search algorithm in R-CNN \cite{girshick2014rich} and Fast R-CNN \cite{girshick2015fast}. The RPN is more computationally efficient and also has a better detection performance.
In Faster R-CNN, the image is fed into a pre-trained CNN to produce a convolutional feature map.
This feature map is then used by the RPN to compute region proposals.
Similar to Fast R-CNN, these region proposals are filtered by ROI pooling and fed into a fully connected layer for classifying objects. 

Region proposals are obtained by sliding a window over the convolutional feature map.
A set $k$ anchor boxes with different scales and ratios are used  at each feature map location.
In total there are $WHk$ anchors for a feature map of height $H$ and Width $W$.
The original implementation of this model uses nine anchor boxes with three different ratios $(1:1, 1:2, 2:1)$  and scales $(128^2, ~256^2, ~512^2)$. Figure~\ref{faster_anchors_default} shows the 9 anchor boxes at point $(300, 500)$ for an image of size $(600,1000)$.
The authors of Faster R-CNN do not mention how the anchor box parameters should be set.
However, they perform an ablation study that varies the number of scales and ratios, indicating that 9 anchor boxes seems to be optimal \cite{ren2016faster}.
Performance of the detector is shown to be sensitive to the input image size, scale and ratio of the anchor boxes used. 
The input image size is defined as the pixel size of an image's shortest side. 

For this detector we consider three sets of hyper-parameters: The input image size, and the anchor box scale and ratios. In order to reduce the number of assumptions, we take all hyper-parameters to be continuous, with a given transformation into the original parameters of Faster R-CNN, as specified by Table~\ref{hyp_cmaes_faster}.

\begin{table}[t]
    \caption{Hyper-parameter Space for Faster R-CNN, with $B_s= 16$.}
    \begin{center}
        \begin{tabular}{lllllll}
            \toprule
            \textbf{Hyper-param} & \textbf{Range} & \textbf{Scaled Value} & \textbf{Type} & \textbf{Transform} & \textbf{Pixel Value} & \\
            \hline\noalign{\smallskip}
            Input Image Size $s_0$ & $(0.3, 0.7{]}$ & 0.6 & C & $1000 s_0$ & 600 &  \\
            \midrule
            Anchor Ratio 1 $r_1$ & $(0, 1{]}$ & 0.25 & C & $2 r_1$ & 0.5 &  \\
            Anchor Ratio 2 $r_2$ & $(0, 1{]}$ & 0.5 & C & $2 r_2$ & 1 &  \\
            Anchor Ratio 3 $r_3$ & $(0, 1{]}$ & 1 & C & $2 r_3$ & 2 &  \\
            \midrule
            Anchor Scale 1 $s_1$ & $(0, 1{]}$ & 0.25 & C & $32 B_s  s_1$ & 128 &  \\
            Anchor Scale 2 $s_2$ & $(0, 1{]}$ & 0.5 & C & $32 B_s s_2$ & 256 &  \\
            Anchor Scale 3 $s_3$ & $(0, 1{]}$ & 1 & C & $32 B_s s_3$ & 512 &  \\
            \bottomrule
        \end{tabular}
    \end{center}
    \label{hyp_cmaes_faster}
\end{table}

\begin{figure}[!t]
    \centering
    \includegraphics[width=1\columnwidth]{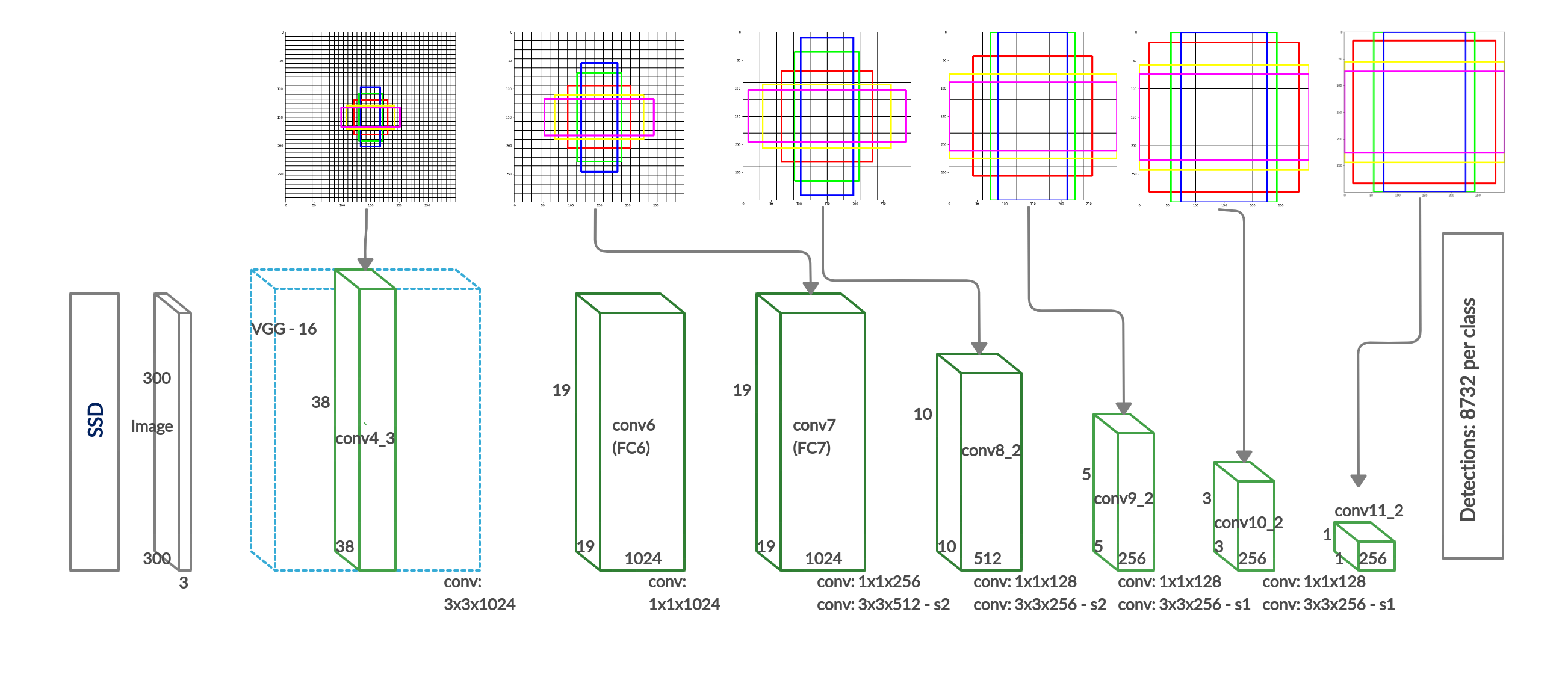}
    \caption{SSD Architecture showing prior boxes at each layer. Image adapted from \cite{liu2016ssd}}
    \label{ssd_colo_anchor}
\end{figure}

\textbf{Single Shot Multi-Box Detector (SSD)} \cite{liu2016ssd} is a single-stage object detector.
This architecture uses feature maps obtained from the backbone network VGG to construct additional convolutional layers corresponding to different anchor scales and ratios. Figure~\ref{ssd_colo_anchor} shows the prior anchor boxes with varying scales and ratios on each convolutional layer.

Each feature map location predicts the offsets relative to the multiple prior anchor boxes along with the confidences for each class.
The loss function is computed by matching the prior anchor boxes with the ground truth boxes using the IOU score.
Hence, designing prior boxes scales and ratios is essential for having better alignments of these prior boxes to each position of the feature map.
The design of prior boxes depends on the size and ratio of objects in the dataset. In SSD, the scale sizes are determined using Equation~\ref{anchor_scale_calc}:
\begin{equation}
    \begin{aligned}
        s_k = s_{\text{min}} + \frac{s_{\text{max}} - s_{\text{min}}}{m-1}(k-1), \, k \in[1,m]
        \label{anchor_scale_calc}
    \end{aligned}
\end{equation}
Where $s_k$ represents the scale size at $k^{\text{th}}$ feature map and $m$ denotes the number of feature maps. The minimum scale size is $s_{\text{min}} = 0.2$, and maximum scale size is $s_{\text{max}} = 0.9$. The conv$4\_3$ prior box scale is fixed to $0.1$. There are four non-square anchor ratios and one square ratio ($a_r \in {1,2,3,\frac{1}{2},\frac{2}{3}}$) for each scale. There is also a constant box with anchor ratio 1 whose scale is computed using $s^{'}_{k} = \sqrt{s_{k}s_{k+1}}$.  The width and height of the default boxes are computed as:
\begin{equation}
    \begin{aligned}
        w_{k}^{a} = s_{k} \sqrt{a_{r}} \qquad
        h_{k}^{a} = \frac{s_{k}}{\sqrt{a_{r}}}.
        \label{anchor_width_height}
    \end{aligned}
\end{equation}
Some feature map skip the $\{\frac{1}{3}, 3\}$ anchor ratios. While training SSD with the MS COCO dataset, the authors reduce the minimum scale and conv$4\_3$ prior box scale to tackle the smaller objects in MS-COCO dataset.

For this detector, we decided to tune the scales $s_i$ and treat them as continuous hyper-parameters, as shown in Table~\ref{hyp_space_ssd}. All other hyper-parameters' values are kept from the original implementation.

\begin{table}[!t]
    \caption{Hyper-parameter Space for the Single-Shot MultiBox Detector (SSD). $a_r$ denotes the anchor ratio.}
    \label{hyp_space_ssd}
    \begin{center}
        \begin{tabular}{llllll}
            \toprule
            \textbf{Hyper-param} & \textbf{Range} & \textbf{Type} & \textbf{Feature map} & \multicolumn{2}{l}{\textbf{Anchor Box}} \\
            &  &  &  & \textbf{Width} $w_k$ & \textbf{Height} $h_k$  \\
            \midrule
            \textbf{Scale 0} ($s_0$) & $(0, 1.06 {]}$ & C & Conv 4\_3 & $s_{0} \sqrt{a_{r}}$ & $\frac{s_{0}}{\sqrt{a_{r}}}$ \\
            \textbf{Scale 1} ($s_1$) & $(0, 1.06{]}$ & C & Conv 7(fc7) & $s_{1} \sqrt{a_{r}}$ & $\frac{s_{1}}{\sqrt{a_{r}}}$ \\
            \textbf{Scale 2} ($s_2$) & $(0, 1.06{]}$ & C & Conv 8\_2 & $s_{2} \sqrt{a_{r}}$ & $\frac{s_{2}}{\sqrt{a_{r}}}$ \\
            \textbf{Scale 3} ($s_3$) & $(0, 1.06{]}$ & C & Conv 9\_2 & $s_{3} \sqrt{a_{r}}$ & $\frac{s_{3}}{\sqrt{a_{r}}}$ \\
            \textbf{Scale 4} ($s_4$) & $(0, 1.06{]}$ & C & Conv 10\_2 & $s_{4} \sqrt{a_{r}}$ & $\frac{s_{4}}{\sqrt{a_{r}}}$ \\
            \textbf{Scale 5} ($s_5$) & $(0, 1.06{]}$ & C & Conv 11\_2 & $s_{5} \sqrt{a_{r}}$ & $\frac{s_{5}}{\sqrt{a_{r}}}$ \\
            \textbf{Scale 6} ($s_6$) & $(0, 1.06{]}$ & C & NA & NA & NA \\
            \bottomrule
        \end{tabular}
    \end{center}            
\end{table}

In the following two sections we show our main experimental results. We evaluate Faster R-CNN and SSD on the PASCAL VOC 2007/2012 and MS COCO datasets, and learn hyper-parameters using CMA-ES and Bayesian Optimization.

\section{Experiments on Faster R-CNN}

The training and validation split of PASCAL VOC 2007 is used for training, and the fitness score (mAP) is computed on the test split of PASCAL VOC 2007. Table 1 in the supplementary shows the CMA-ES parameters used in this experiment. Training and evaluation of Faster R-CNN for all hyper-parameters of a particular generation were done in parallel. We use the Faster Pytorch Implementation \footnote{\url{https://github.com/jwyang/faster-rcnn.pytorch}} \cite{jjfaster2rcnn} of Faster R-CNN.

\begin{figure}[t]
    \centering
    \includegraphics[width=0.5\columnwidth]{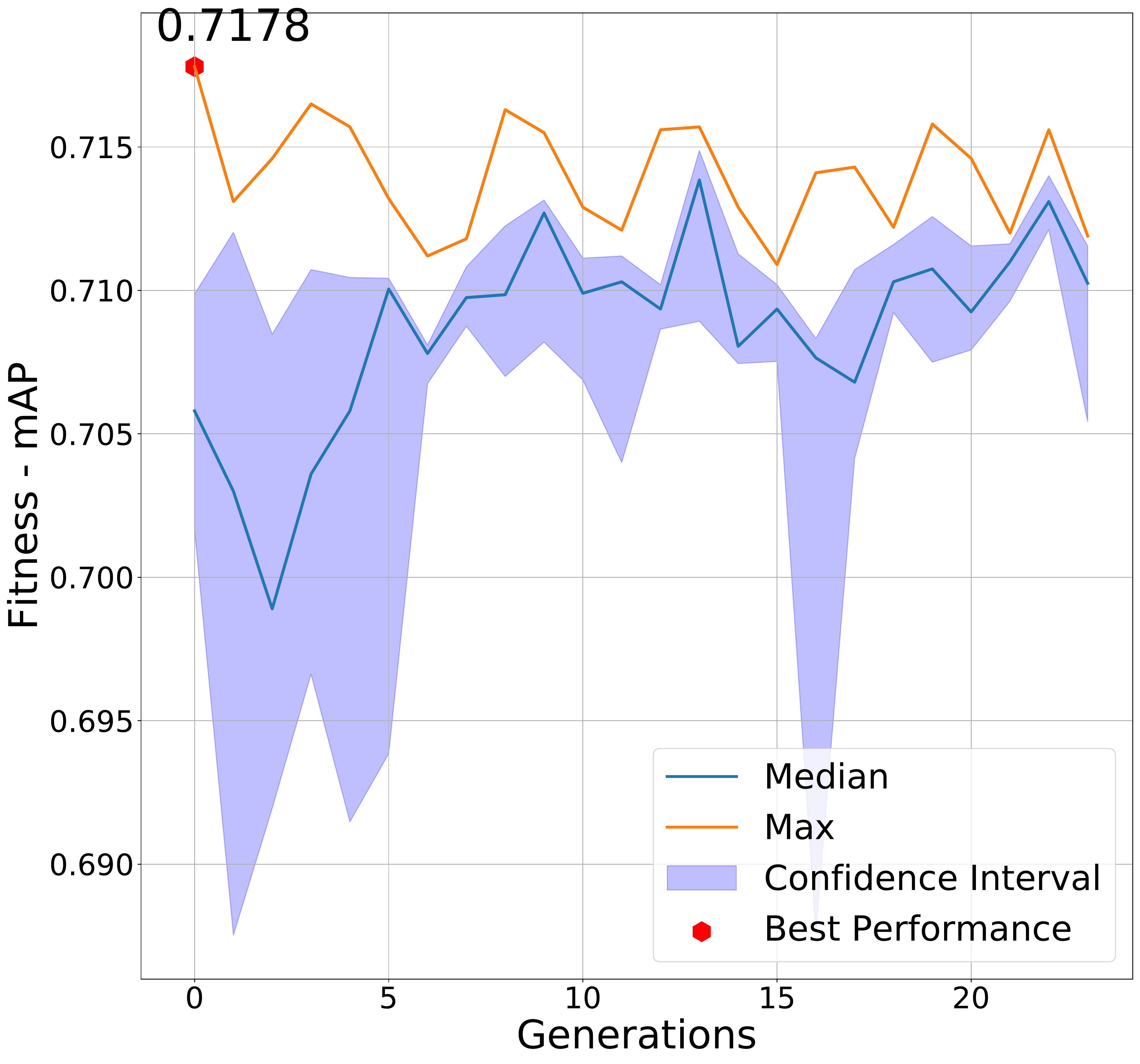}
    \caption{Faster R-CNN -Optimization by CMA-ES on PASCAL VOC 2007}
    \label{faster_results_cmaes_voc}
\end{figure}

\begin{figure}[!tb]
    \centering
    \begin{subfigure}[b]{0.45\textwidth}
        \includegraphics[width=\textwidth]{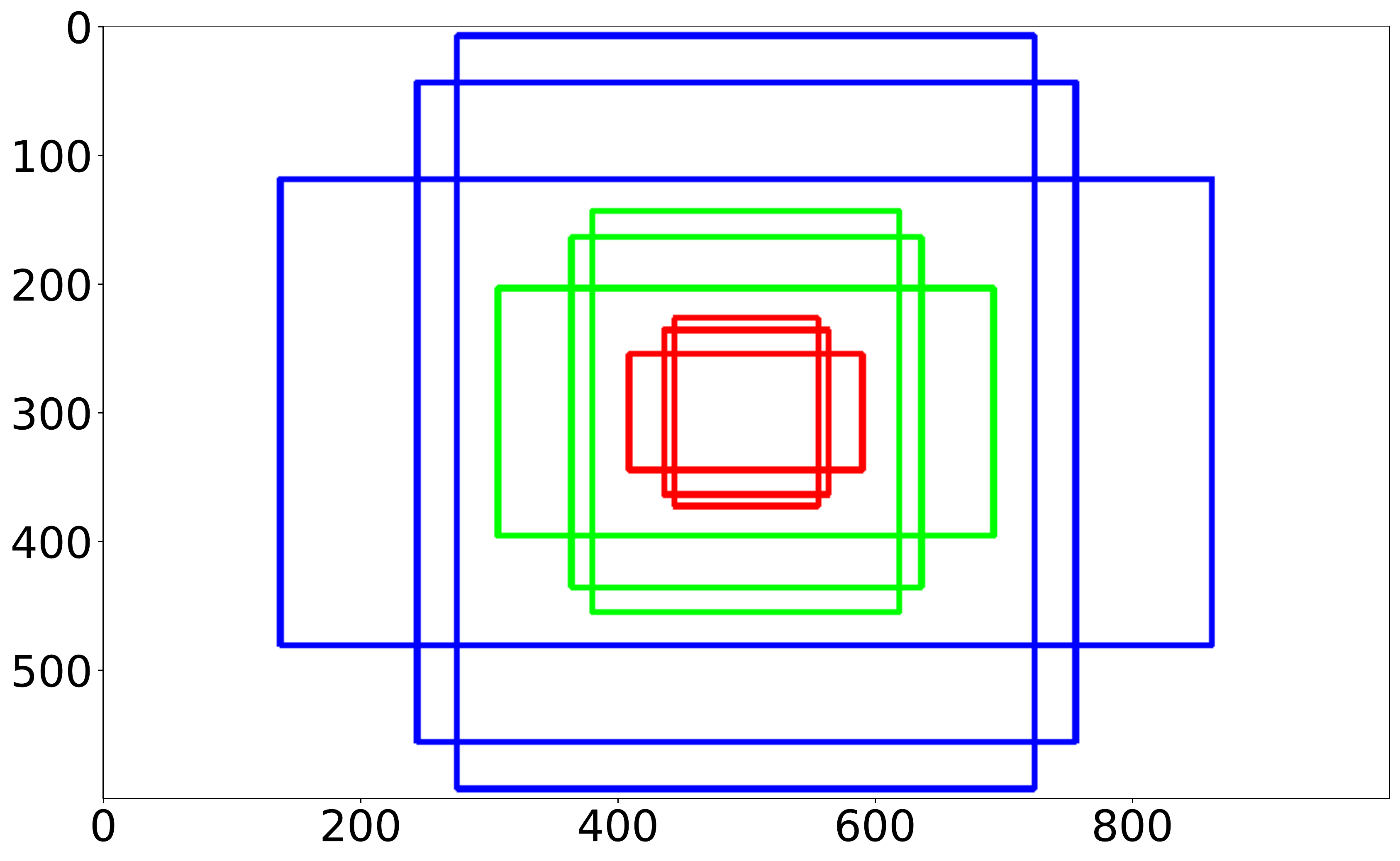}
        \caption{Original prior anchor boxes. Image scale = 600}
        \label{faster_anchors_default}
    \end{subfigure}
    ~
    \begin{subfigure}[b]{0.45\textwidth}
        \includegraphics[width=0.85\textwidth]{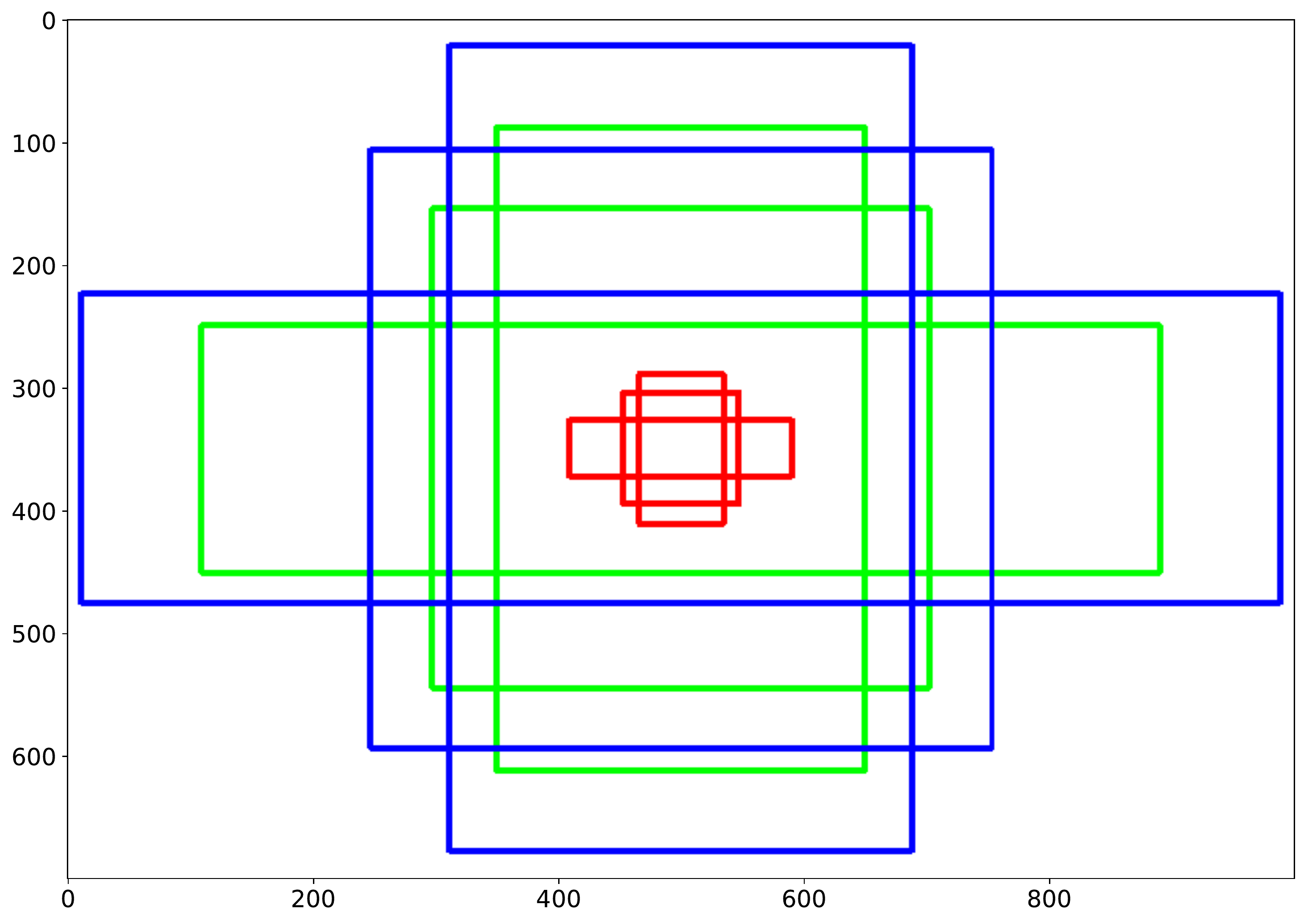}
        \caption{Prior anchor boxes found using CMA-ES. Image scale = 699}
        \label{faster_anchors_cmaes}
    \end{subfigure}
    \caption{Faster R-CNN prior anchor boxes comparison for PASCAL VOC 2007}
    \label{faster_anchors_overall}
\end{figure}

Figure \ref{faster_results_cmaes_voc} shows the performance of Faster R-CNN optimized using CMA-ES for 25 generations (225 evaluations) on PASCAL VOC 2007 dataset. This setup achieves an mAP of 71.78 \% which is 1.79\% more than the default hyper-parameters (69.9\% mAP). Figure \ref{faster_anchors_overall} shows the comparison of prior boxes in the original implementation of Faster R-CNN with the prior boxes found using CMA-ES. Table \ref{tab:faster_res_sum} shows an overall comparison of all our results along with the associated anchor scales and aspect ratios.

\subsection{Regression Analysis}
\label{reg_analysis_frcnn}

An important question is to determine which are the best scales to tune, as some scales might be more important than others. To answer this question, we performed regression analysis between the object detector performance, as measured by mAP, and the individual hyper-parameters. Regression analysis can explain the relationship between a dependent variable and many independent variables, and can also explain the significance of the independent variables.

For this we normalize the mAP and all hyper-parameters to the $[0, 1]$ range, and trained a linear regression model. The coefficients of this model can then be interpreted as importance scores, which should tell us about which hyper-parameter is more important. We measure the goodness of fit using the coefficient of determination $R^2$, which indicates the amount of variance explained by the independent variable (the mAP).

Table \ref{reg_faster} shows results of the regression analysis with Faster R-CNN hyper-parameters.
The higher coefficient value of image scale indicates that tuning the image scales in Faster R-CNN is very important, but the highest coefficient is associated to the input image size ($s_0$), indicating that it has the highest effect on mAP. Anchor Scales one and two have a significant coefficient value, indicating their importante, which we believe makes sense as they have the biggest receptive field size. Adjusting the ratios ($r_1-r_3$) seems to have little impact on the overall mAP.

\begin{table}[!t]
    \caption{Regression Analysis of Faster R-CNN Hyper-parameters}
    \label{reg_faster}
    \begin{center}
        \begin{tabular}{lllllllll}
            \toprule
            \textbf{Method} & $\mathbf{R^2}$ & \multicolumn{7}{c}{\textbf{Coefficients}} \\
            \cmidrule{3-9} 
            & \multicolumn{1}{c}{} & \multicolumn{1}{c}{\textbf{\begin{tabular}[c]{@{}c@{}}Input \\ Size $(s_0)$\end{tabular}}} & \multicolumn{1}{c}{\textbf{\begin{tabular}[c]{@{}c@{}}Scale\\ 1 $(s_1)$\end{tabular}}} & \multicolumn{1}{c}{\textbf{\begin{tabular}[c]{@{}c@{}}Scale\\ 2 $(s_2)$\end{tabular}}} & \multicolumn{1}{c}{\textbf{\begin{tabular}[c]{@{}c@{}}Scale\\ 3 $(s_3)$\end{tabular}}} & \multicolumn{1}{c}{\textbf{\begin{tabular}[c]{@{}c@{}}Ratio\\ 1 $(r_1)$\end{tabular}}} & \multicolumn{1}{c}{\textbf{\begin{tabular}[c]{@{}c@{}}Ratio\\ 2 $(r_2)$\end{tabular}}} & \multicolumn{1}{c}{\textbf{\begin{tabular}[c]{@{}c@{}}Ratio\\ 3 $(r_3)$\end{tabular}}} \\
            \midrule
            CMA-ES & 0.52 & 0.67 & 0.25 & 0.19 & 0.02 & 0.02 & 0.01 & 0.07 \\
            \begin{tabular}[c]{@{}l@{}}CMA-ES\\ (Without $s_0$)\end{tabular} & 0.21 &  & 0.32 & 0.15 & 0.01 & 0.01 & 0.02 & 0.08 \\
            \bottomrule
        \end{tabular}
    \end{center}
\end{table}    

\begin{table}[!b]
    \caption{Summary of our Faster R-CNN results on PASCAL VOC test 2007.}
    \label{tab:faster_res_sum}        
    \begin{center}          
        \begin{tabular}{llll}
            \toprule
            \textbf{Settings} & \textbf{Anchor Scales} & \textbf{Aspect Ratios} & \textbf{mAP(\%)} \\ 
            \midrule
            \textbf{Default} & $128^2 , 256^2, 512^2$ & 0.5, 1, 2 & 69.9 \\ 
            \textbf{BOGP} & $48^2 , 144^2, 512^2$ & 0.941 , 1.155, 2.015 & 71.37 \\ 
            \textbf{SMAC} & $80^2 , 304^2, 512^2$ & 0.4, 0.5, 1 & 71.56 \\ 
            \bottomrule
            \textbf{CMA-ES} & $92^2 , 397^2, 497^2$ & 0.259,  0.964, 1.741 & 71.78 \\ \hline
        \end{tabular}%
    \end{center}
\end{table}

\section{Experiments on SSD}

\subsection{Prior Boxes Optimization using CMA-ES}

Quality of optimization using CMA-ES highly depends on setting proper initial parameters. We set the population size $\lambda$ for PASCAL VOC 2007 based on $\lambda = 4+3\ln(n)$ \cite{hansen2016cma}. For MS-COCO, we use a smaller population size $\lambda = 4$ based on our GPU resource availability. CMA-ES parameters that we use are shown in Table 2 in the supplementary. We set the initial scales according to Equation \ref{anchor_scale_calc}, with values ${[}0.1,  0.2,  0.37,  0.54,  0.71,  0.88,  1.05{]}$.  Mean average precision (mAP) is used as a fitness score for the individuals.  We use a high quality implementation \cite{lufficc2018ssd} of SSD\footnote{\url{https://github.com/lufficc/SSD}} in PyTorch.

\begin{figure}[!tb]
    \begin{subfigure}[b]{0.45\textwidth}
        \includegraphics[width=\textwidth]{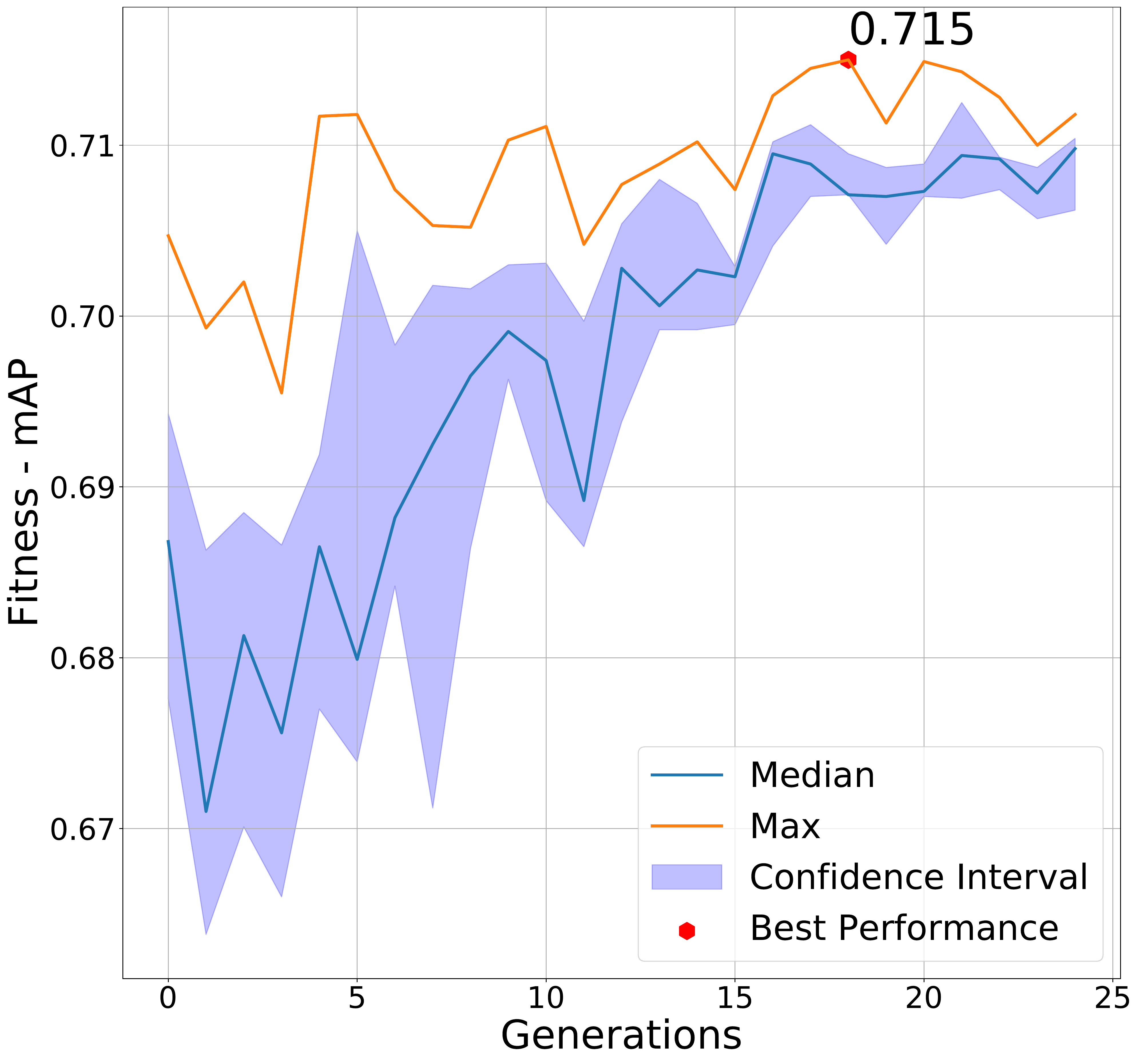}
        \caption{PASCAL VOC 2007}
        \label{cmaes_fitness_plot_voc}
    \end{subfigure}
    \begin{subfigure}[b]{0.45\textwidth}
        \includegraphics[width=\textwidth]{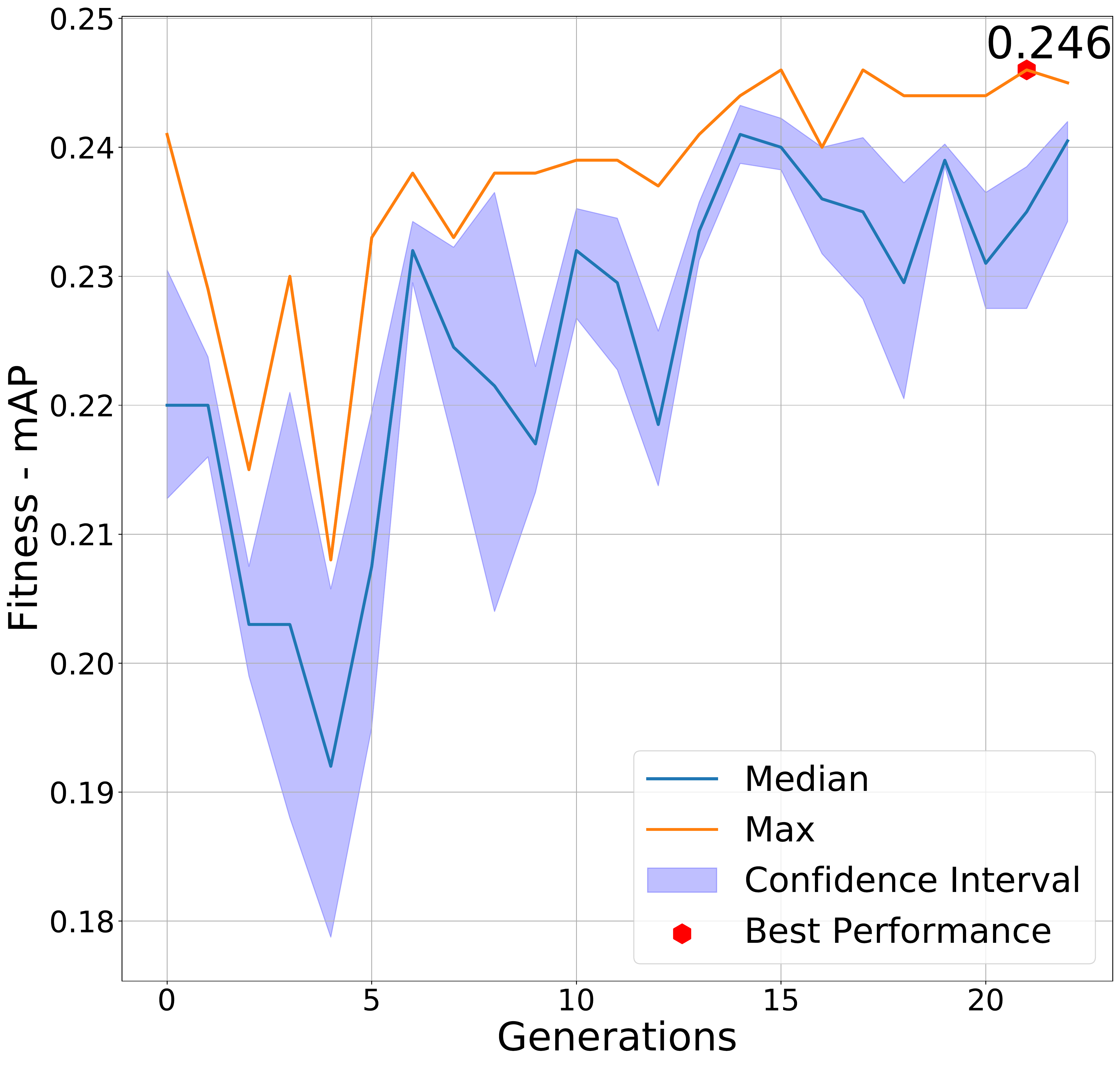}
        \caption{MS COCO}
        \label{cmaes_fitness_plot_coco}
    \end{subfigure}
    \caption{Optimizing SSD hyperparameters using CMA-ES.}
    \label{cmaes_fitness_plot}
\end{figure}

\textbf{Anchor Scales on PASCAL VOC 2007}. We use the training and validation split of PASCAL VOC 2007 for training, and the fitness score (mAP) is computed on the test split. We evaluate hyper-parameter configurations by training SSD on nine different Nvidia TitanXP GPUs. The training and evaluation of a single hyper-parameter configuration on SSD takes around 24.5 hours. Hence, the evaluation of one generation can be completed in a single day if it is done in parallel.

\begin{figure}[t]
    \centering
    \includegraphics[width=1\columnwidth]{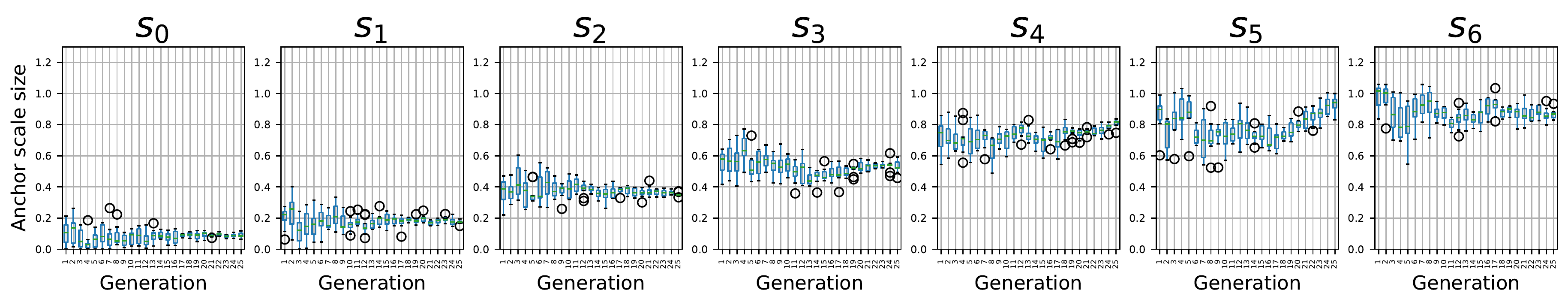}
    \caption{Box plot of SSD Anchor Scale values over generations on PASCAL VOC 2007}
    \label{anchor_cmaes_box_07}
\end{figure}

Figure \ref{cmaes_fitness_plot_voc} shows the performance of SSD optimized using CMA-ES for 25 generations (225 evaluations). We can clearly see performance improvements over generations. The median mAP also increases with each generation, depicting the auto-tuning of anchor scales. Figure \ref{anchor_cmaes_box_07} shows the box plot of each anchor scale values over generation, and it illustrates the convergence of each anchor scale towards the best value. The best anchor scale configuration found using CMA-ES has achieved a mean average precision of 71.55\%, which is 3.55\% higher than the default scales (68.0 \% mAP).

\textbf{Anchor Scales on MS COCO}.	We also evalaute on MS COCO using CMA-ES. For this dataset, the authors of SSD reduce the minimum scale and $conv 4\_3$ prior box scale to tackle the smaller objects in the MS-COCO dataset as $0.1$5 and $0.07$ respectively.
Instead of using the modified scales as an initial scales for CMA-ES, we used the regularly placed scales according to Equation \ref{anchor_scale_calc}, and the constant box scale of $0.1$ to analyze the performance of CMA-ES for a dataset with a lot of smaller objects.
We use the \texttt{trainval35k} split of MS COCO 2014 for training, and the fitness score (mAP) is computed on the \texttt{minival} split. The training and evaluation of a single hyper-parameter configuration took around 44 hours.

Figure \ref{cmaes_fitness_plot_voc} shows the performance of SSD optimized using CMA-ES for 23 generations (92 evaluations). We clearly see the increase in fitness over generations by tuning the anchor scales using CMA-ES. The anchor scales found using CMA-ES adapted well as it is able to find smaller scales, appropriate for a dataset with many small objects. However, the anchor scales found by CMA-ES achieve 24.6\% mAP$@[0.5:0.95]$ which is 0.5\% less in comparison with the original anchor scales. The reason for this is apparently we used a small population size due to GPU resource constraints. With a higher population size, CMA-ES should be able to find a better anchor scale as it had proved to optimize the scales with an increase in generations with even small population size. Table \ref{cmaes_res_map} shows the detailed mAP results, with improvements in AP for medium and large object sizes

\begin{table}[!htb]
    \begin{center}
        \begin{tabular}{lllllllll}
            \toprule
            \multicolumn{1}{l}{\textbf{Settings}} & & \multicolumn{3}{l}{\textbf{Avg. Precision (AP)}} & & \multicolumn{3}{l}{\textbf{Avg. Precision (AP)}} \\ 
            \cmidrule{3-5} \cmidrule{7-9}
            \multicolumn{1}{l}{} & IOU & \multicolumn{1}{l}{0.5:0.95} & \multicolumn{1}{l}{0.5} & \multicolumn{1}{l}{0.75} & Area & \multicolumn{1}{l}{\textbf{Small}} & \multicolumn{1}{l}{\textbf{Medium}} & \multicolumn{1}{l}{\textbf{Large}} \\
            \midrule
            \textbf{Default} & & 25.1 & 43.1 & 25.8 & & 6.6 & 25.9 & 41.4 \\ 
            \textbf{CMA-ES} & & 24.6 & 42.8 & 25.1 & & 4.8 & 26.3 & 42.9 \\
            \bottomrule
        \end{tabular}
    \end{center}
    \caption{Comparison of SSD Scales on MS COCO - CMA-ES vs Default Scales}
    \label{cmaes_res_map}
\end{table}

\subsection{Prior Boxes Optimization using Bayesian Optimization}	
We also experiment with Bayesian Optimization to tune the prior anchor box scales of SSD for PASCAL VOC 2007. We set the same hyper-parameter space of the anchor scale as shown in  Table \ref{hyp_space_ssd}. Expected Improvement (EI) is used as the acquisition function in both BOGP and SMAC.
The experiment was carried out sequentially with computation budget of 75 function evaluations. Our results are presented in Figure \ref {ssd_colo_anchor}. Though both BOGP and SMAC achieve almost the same result best mAP, BOGP took less function evaluations when compared to both SMAC and CMA-ES. 
However, after finding the best scales, the performance of many anchor scales generated by BOGP are not satisfactory, which is indicated by the plot scattering. This is because the optimization algorithm might try to increase its exploration space in search of better hyper-parameters.

A summary of our results is shown in Table \ref{tab:ssd_results_sum}, including the associated anchor scales. We include a baseline where we obtained anchor scales using k-means on the training set, similar to YOLOv2 \cite{redmon2017yolo9000}. There was only a small improvement by using k-means scales, and it is outperformed by all hyper-parameter tuning methods.

\begin{figure}[t]
    \centering
    \includegraphics[width=0.75\columnwidth]{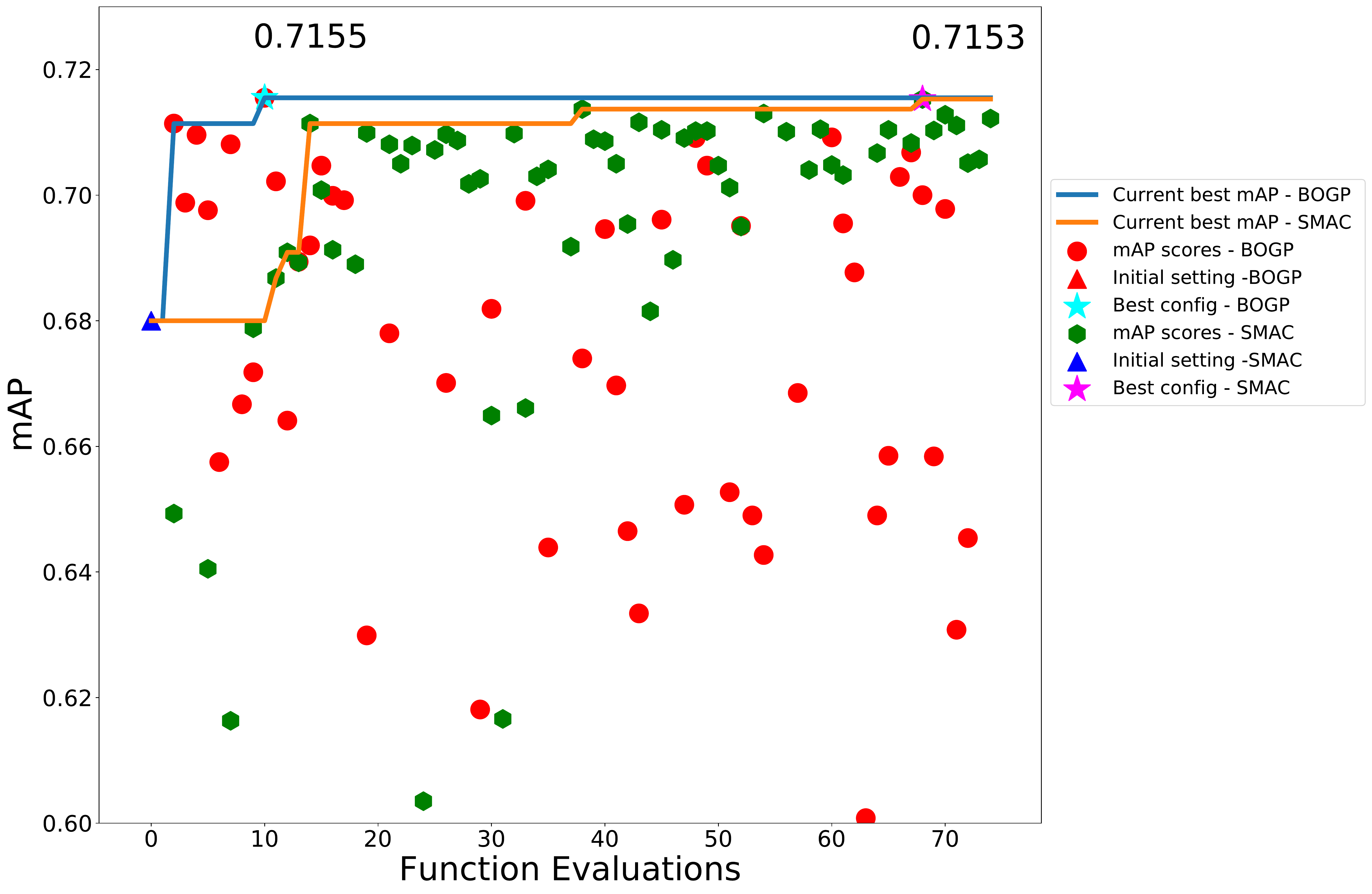}
    \caption{Optimizing SSD hyper-parameters using Bayesian Optimization and SMAC on PASCAL VOC 2007}
    \label{ssd_colo_anchor}
\end{figure}

\begin{table}[!htb]
    \caption{Summary of our SSD results on PASCAL VOC test 2007.}
    \label{tab:ssd_results_sum}
    \begin{center}
        \begin{tabular}{lllllllll}
            \toprule
            \textbf{Method} &  \multicolumn{7}{c}{\textbf{Anchor Scales}} & \textbf{mAP} \\ \cmidrule{2-8} 
            &   \multicolumn{1}{c}{\textbf{\begin{tabular}[c]{@{}l@{}}Scale \\ 0\end{tabular}}} & \multicolumn{1}{l}{\textbf{\begin{tabular}[c]{@{}l@{}}Scale \\ 1\end{tabular}}} & \multicolumn{1}{l}{\textbf{\begin{tabular}[c]{@{}l@{}}Scale \\ 2\end{tabular}}} & \multicolumn{1}{l}{\textbf{\begin{tabular}[c]{@{}l@{}}Scale \\ 3\end{tabular}}} & \multicolumn{1}{l}{\textbf{\begin{tabular}[c]{@{}l@{}}Scale \\ 4\end{tabular}}} & \multicolumn{1}{l}{\textbf{\begin{tabular}[c]{@{}l@{}}Scale \\ 5\end{tabular}}} & \multicolumn{1}{l}{\textbf{\begin{tabular}[c]{@{}l@{}}Scale\\  6\end{tabular}}} &
            \multicolumn{1}{l}{\textbf{\begin{tabular}[c]{@{}l@{}}\end{tabular}}} \\
            \midrule
            \textbf{Default} & 0.1&  0.2&  0.37&  0.54&  0.71&  0.88& 1.05 & 68.0 \\
            \midrule
            \textbf{BOGP} & 0.0973& 0.2132& 0.3689& 0.5224& 0.5882& 1.038& 1.051 & 71.55 \\ 
            \textbf{SMAC} & 0.1037& 0.1935& 0.3707& 0.5542& 0.6934& 0.9481& 1.0481 & 71.53 \\ 
            \textbf{CMA-ES} & 0.0908& 0.1892& 0.3276&  0.5240& 0.6865&  0.8390&  0.8894 & 71.50 \\
            \textbf{k-Means} & N/A & N/A & N/A & N/A & N/A & N/A & N/A & 68.35\\
            \bottomrule
        \end{tabular}
    \end{center}
\end{table}

\subsection{Generalization Analysis}   

So far we have optimized object detector scales based on its evaluation score on PASCAL VOC 2007 test dataset. Hyper-parameter optimization may also overfit to the evaluation dataset \cite{cawley2010over}. To verify the generalization of the learned scales, we tested the scales learned on PASCAL VOC 2007 on a different dataset, namely the PASCAL VOC 2012 validation set. As seen in Table \ref{perf_comp_gen}, it is evident that the optimized models still able to perform better than the hand-tuned scales on the PASCAL VOC 2012 validation set. Table \ref{perf_comp_gen} indicates that SMAC has improved generalized performance in comparison with other methods.

\begin{table}[!htb]
    \caption{Performance comparison of optimized SSD scales on PASCAL VOC test 2007 and PASCAL VOC validation 2012}
    \label{perf_comp_gen}
    \begin{center}
        \begin{tabular}{llllll}
            \toprule
            \textbf{Dataset} & \textbf{Default} & \textbf{BO-GP} & \textbf{SMAC} & \textbf{CMA-ES} & \textbf{k-Means}\\ 
            \midrule
            \textbf{VOC 2007 Test} & 68.0 & \textbf{71.55} & 71.53 & 71.50 & 68.35\\ 
            \textbf{VOC 2012 Val} & 61.74 & 63.54 & \textbf{64.43} & 63.63 & 61.91\\ 
            \bottomrule
        \end{tabular}
    \end{center}              
\end{table}

\subsection{Regression Analysis}
\label{reg_analysis_ssd}

Similarly to Section \ref{reg_analysis_frcnn}, we performed regression analysis to find the importance of each anchor scale. We used the absolute of the coefficient to find the most important hyper-parameter. Table \ref{reg_ssd} shows our results. In BO and SMAC, we can see anchor scale zero and one have the largest coefficients. Moreover, both BOGP and SMAC agrees that anchor scale 0 is a very important hyper-parameter, and the high $R^2$ score also indicates a good fit, validating that this scale is probably the most important to tune.

The higher $R^2$ score of regression from SMAC results indicates a good explanation of the mAP given the anchor scales. In CMA-ES, anchor scales two, three, and four have the largest coefficients, but the low $R^2$ score makes this relation not significant.

It is interesting that the explainability of the mAP given the scales varies considerably with the hyper-parameter optimization methods. One would imagine that there should be no such relation, but each method guides sampling on the hyper-parameter space in a different way, which might introduce certain biases.

\begin{table}[t]
    \caption{Regression analysis with SSD hyper-parameters on PASCAL VOC 2007}
    \label{reg_ssd}
    \begin{center}
        \begin{tabular}{lllllllll}
            \toprule
            \textbf{Method} & $\mathbf{R^2}$ & \multicolumn{7}{c}{\textbf{Coefficients}} \\ \cmidrule{3-9} 
            &  & \multicolumn{1}{c}{\textbf{\begin{tabular}[c]{@{}l@{}}Scale \\ 0\end{tabular}}} & \multicolumn{1}{l}{\textbf{\begin{tabular}[c]{@{}l@{}}Scale \\ 1\end{tabular}}} & \multicolumn{1}{l}{\textbf{\begin{tabular}[c]{@{}l@{}}Scale \\ 2\end{tabular}}} & \multicolumn{1}{l}{\textbf{\begin{tabular}[c]{@{}l@{}}Scale \\ 3\end{tabular}}} & \multicolumn{1}{l}{\textbf{\begin{tabular}[c]{@{}l@{}}Scale \\ 4\end{tabular}}} & \multicolumn{1}{l}{\textbf{\begin{tabular}[c]{@{}l@{}}Scale \\ 5\end{tabular}}} & \multicolumn{1}{l}{\textbf{\begin{tabular}[c]{@{}l@{}}Scale\\  6\end{tabular}}} \\
            \midrule
            CMA-ES 	& 0.15 & 0.15 & 0.04 & 0.38 & 0.50 & 0.28 & 0.03 & 0.05 \\
            BOGP 	& 0.66 & 0.54 & 0.31 & 0.12 & 0.04 & 0.02 & 0.00 & 0.22 \\
            SMAC 	& 0.82 & 0.82 & 0.17 & 0.02 & 0.06 & 0.12 & 0.09 & 0.08 \\
            \bottomrule
        \end{tabular}
    \end{center}		
\end{table}

\section{Conclusions and Future Work}

In this paper, we demonstrate the performance of Black-box optimization for object detection hyper-parameters, in particular the default box/anchor scales, on PASCAL VOC 2007 and MS COCO. From our experimental results, we can conclude that using Black-box optimization produces an improvement on mAP by adjusting the anchor/prior box scales of Faster R-CNN and SSD on the PASCAL VOC 2007 and MS-COCO datasets, and generally can achieve better results than the hand-tuned configurations in most of the cases. In MS-COCO we observed decreased performance for small objects, and only medium and large objects see an improvement in mAP.

GP-based Bayesian Optimization obtains better performance with less function evaluations. CMA-ES results show a clear view of the improvement in performance with an increasing number of generations. BOGP and SMAC can be studied more in with different initial designs.

We also evaluated the transferability of the learned scales with SSD on PASCAL VOC 2007, by evaluating these scales on the PASCAL VOC 2012 validation set, which also shows an improvement in mAP. We believe this shows that learning scales using these methods is a good alternative to manually designing the scales by a human. It is possible that the scales currently used by object detectors are sub-optimal, in terms that each dataset has probably a different set of optimal scales. Using automatic tuning methods should help practitioners tune an object detector for a particular dataset with minimal effort.

Finally, for each combination we performed a simple regression analysis to find out the importance of each hyper-parameter to the overall mAP. We find that for Faster R-CNN, the biggest factor is the input image size, while for SSD the first scales contribute more to increasing mAP. This information is valuable for future research, as efforts can be concentrated on the most important hyper-parameters, decreasing search time and computational costs.

Broader research is needed on larger hyper-parameter spaces by including other object detection hyper-parameters that were not used in our experiments, for example, it would be very interesting to tune not only the actual scales, but also the number of scales, which may vary with the dataset, in particular with MS COCO as it contains many small objects. The multi-task loss weights are also not generally tuned, which could be a possible source of improvement. Multi-objective optimization is can be used to choose the hyper-parameters that minimize prediction time while maximizing the task performance (higher mAP) of the object detector.


\clearpage
%
%
\bibliographystyle{splncs}
\bibliography{egbib}

\appendix
\clearpage
\section{Parameters of Optimization Methods}

In this section we provide details of the parameters of CMA-ES, as shown in Table \ref{cmaes_faster_parameters} and Table \ref{cmaes_ssd_parameters}.

\begin{table}[!h]
    \caption{CMA-ES parameters used in this experiments with Faster R-CNN and initial vector {[}0.6 ,0.25, 0.5, 1.0, 0.25, 0.5, 1.0{]}}
    \label{cmaes_faster_parameters}
    \begin{center}
        \begin{tabular}{lllllll}
            \toprule
            \textbf{Dataset} & \textbf{\begin{tabular}[c]{@{}l@{}}Step Control\\ $\sigma$\end{tabular}} & \textbf{\begin{tabular}[c]{@{}l@{}}Population \\ Size\\ $\lambda$\end{tabular}} &
            \textbf{\begin{tabular}[c]{@{}l@{}}Maximum \\ Evaluations\end{tabular}} &
            \textbf{\begin{tabular}[c]{@{}l@{}}Number\\  of \\ Genes\end{tabular}} & \textbf{\begin{tabular}[c]{@{}l@{}}Number\\ of \\ Generations\end{tabular}} \\
            \midrule
            \begin{tabular}[c]{@{}l@{}}
                PASCAL \\ VOC 2007\end{tabular} &  0.3 & 6 & 150 & 7 & 25 \\
            \bottomrule
        \end{tabular}
    \end{center}   
\end{table}

\begin{table}[!h]
    \caption{CMA-ES parameters used for SSD experiments. The initial vector is set to {[}0.1, 0.2, 0.37, 0.54, 0.71, 0.88, 1.05{]}.}
    \label{cmaes_ssd_parameters}
    \begin{center}
        \begin{tabular}{lllllll}
            \toprule
            \textbf{Dataset} & \textbf{\begin{tabular}[c]{@{}l@{}}Step\\ control\\ $\sigma$\end{tabular}} & \textbf{\begin{tabular}[c]{@{}l@{}}Population \\ Size\\ $\lambda$\end{tabular}} & \textbf{\begin{tabular}[c]{@{}l@{}}Maximum\\  Evaluations\end{tabular}} & \textbf{\begin{tabular}[c]{@{}l@{}}Number of \\ Genes\end{tabular}} & \textbf{\begin{tabular}[c]{@{}l@{}}Number of\\ Generations\end{tabular}} \\
            \midrule
            \begin{tabular}[c]{@{}l@{}}
                PASCAL \\ VOC 2007\end{tabular} & 0.3 & 9 & 225 & 7 & 25 \\
            MS-COCO & 0.3 & 4 & 92 & 7 & 23 \\
            \bottomrule
        \end{tabular}
    \end{center}		
\end{table}

\section{Additional Details of PASCAL VOC Results}

In this section we provide details on the per-class average precision for SSD and Faster R-CNN with different hyper-parameter optimization methods. These results are available in Table \ref{tab_map_ind_ssd_07} for SSD, and in Table \ref{tab_map_ind_faster_07} for Faster R-CNN.

In both cases we see modest improvements in class AP over the default hyper-parameters, up to 4\% in absolute, for example, for the \textit{horse} class in Faster R-CNN.

\begin{table}[!hb]
    \begin{center}
        \begin{tabular}{llllll}
            \toprule
            \textbf{\begin{tabular}[c]{@{}l@{}}Method /\\ Objects\end{tabular}} & \textbf{Default} & \textbf{BOGP} & \textbf{SMAC} & \textbf{CMA-ES} & \textbf{k-Means} \\
            \midrule
            \textbf{aero} & 73.4 & 75.51 & \textbf{76.03} & 74.4	& 76.41 \\
            \textbf{bike} & 77.5 & \textbf{80.17} & 79.19 & 78.86 	& 75.65\\
            \textbf{bird} & 64.1 & 67.24 & \textbf{69.13} & 67.24 	& 63.92\\
            \textbf{boat} & 59.0 & 62.69 & \textbf{66.49} & 63.80 	& 62.46\\
            \textbf{bottle} & 38.9 & 41.24 & \textbf{41.88} & 41.45 & 39.13\\
            \textbf{bus} & 75.2 & \textbf{82.72} & 80.02 & 79.20 	& 76.19\\
            \textbf{car} & 80.8 & \textbf{83.51} & 83.21 & 83.07 	& 82.03\\
            \textbf{cat} & 78.5 & 82.76 & 81.56 & \textbf{84.28} 	& 78.36\\
            \textbf{chair} & 46.0 & 52.48 & \textbf{53.18} & 52.13 	& 48.82\\
            \textbf{cow} & 67.8 & 76.78 & \textbf{77.37} & 76.28 	& 74.36\\
            \textbf{dining table} & 69.2 & 69.50 & \textbf{70.50} & 69.02 & 63.32\\
            \textbf{dog} & 76.6 & 78.66 & 79.31 & \textbf{79.64} 	& 77.28\\
            \textbf{horse} & 82.1 & 81.91 & 82.83 & \textbf{83.74} 	& 78.70\\
            \textbf{motor bike} & 77.0 & \textbf{80.28} & 80.17 & 80.11 & 76.10\\
            \textbf{person} & 72.5 & \textbf{75.37} & 74.91 & 74.98 	& 69.83\\
            \textbf{potted plant} & 41.2 & \textbf{44.62} & 43.57 & 44.47 	& 41.42\\
            \textbf{sheep} & 64.2 & 68.05 & 66.97 & \textbf{70.84} 	& 66.48\\
            \textbf{sofa} & 69.1 & \textbf{73.67} & 71.49 & 71.45 	& 70.09\\
            \textbf{train} & 78.0 & 83.34 & 82.37 & \textbf{83.95} 	& 77.39\\
            \textbf{monitor} & 68.5.5 & \textbf{70.56} & 70.45 & 71.03 	& 69.03\\
            \midrule
            \textbf{mAP} & 68.0 & \textbf{71.55} & 71.37 & 71.50 & 68.35\\
            \bottomrule
        \end{tabular}%
    \end{center}
    \caption[Average precision results comparison - SSD on PASCAL VOC 2007.]{Performance comparison of optimized SSD object detector's average precision of objects  by different blackbox methods on PASCAL VOC test 2007}
    \label{tab_map_ind_ssd_07}
\end{table}

\begin{table}[!htb]
    \begin{center}
        \begin{tabular}{llll}
            \toprule
            \textbf{\begin{tabular}[c]{@{}l@{}}Method  \textbackslash\\ Objects\end{tabular}} & \textbf{Default} & \textbf{BOGP} & \textbf{SMAC} \\
            \midrule
            \textbf{aero} & 70.00 & \textbf{73.71} & 71.6 \\
            \textbf{bike} & 80.6 & \textbf{80.66} & 79.4 \\
            \textbf{bird} & \textbf{70.1} & 69.91 & 68.5 \\
            \textbf{boat} & 57.3 & 56.19 & \textbf{56.90} \\
            \textbf{bottle} & 49.9 & \textbf{57.11} & 57.06 \\
            \textbf{bus} & 78.2 & 76.94 & \textbf{80.88} \\
            \textbf{car} & 80.4 & 82.94 & \textbf{84.42} \\
            \textbf{cat} & 82.0 & 82.46 & \textbf{83.61} \\
            \textbf{chair} & \textbf{52.2} & 50.75 & 50.85 \\
            \textbf{cow} & 75.3 & \textbf{79.96} & 78.63 \\
            \textbf{dining table} & 67.2 & 70.1 & \textbf{70.47} \\
            \textbf{dog} & 80.3 & 79.27 & \textbf{80.65} \\
            \textbf{horse} & 79.8 & \textbf{83.21} & 82.56 \\
            \textbf{motor bike} & 75.0 & 75.48 & \textbf{76.69} \\
            \textbf{person} & 76.3 & \textbf{77.27} & 77.01 \\
            \textbf{potted plant} & 39.1 & 43.23 & \textbf{44.38} \\
            \textbf{sheep} & 68.3 & \textbf{71.33} & 71.22 \\
            \textbf{sofa} & 67.3 & \textbf{67.34} & 66.71 \\
            \textbf{train} & \textbf{81.1} & 77.17 & 76.59 \\
            \textbf{monitor} & 67.6 & 72.45 & \textbf{73.13} \\
            \midrule
            \textbf{mAP} & 69.9 & 71.37 & \textbf{71.56} \\
            \bottomrule
        \end{tabular}%
    \end{center}
    \caption[Average precision results comparison - Faster R-CNN on PASCAL VOC 2007.]{Performance comparison of optimized Faster R-CNN object detector's average precision of objects  by different blackbox methods on PASCAL VOC test 2007}
    \label{tab_map_ind_faster_07}
\end{table}

We also provide details on the anchor scales that we obtained using automatic machine learning.

\begin{figure}[!htb]
    \begin{subfigure}[b]{0.3\textwidth}
        \includegraphics[width=\textwidth]{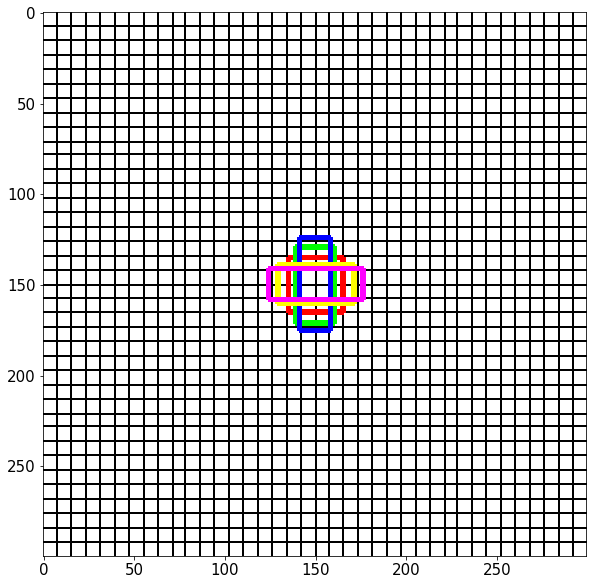}
        \caption{Prior anchor boxes for $38 \times 38$ feature map at conv 4\_3}
        \label{ssd_anchor_plot_default_0}
    \end{subfigure}
    ~
    \begin{subfigure}[b]{0.3\textwidth}
        \includegraphics[width=\textwidth]{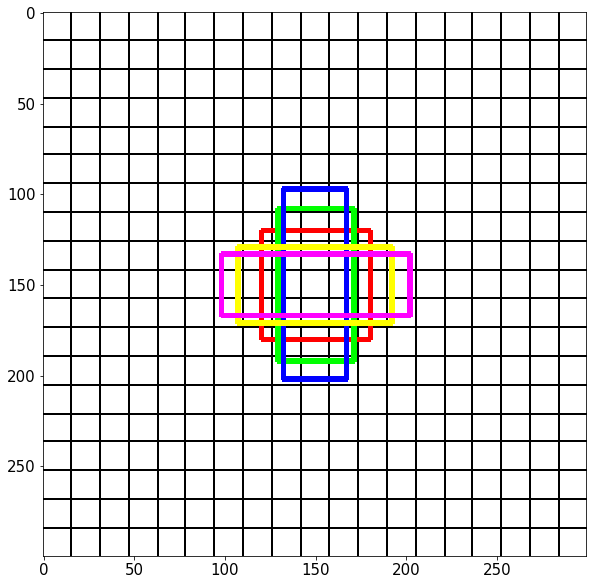}
        \caption{Prior anchor boxes for $19 \times 19$ feature map at conv 7}
        \label{ssd_anchor_plot_default_1}
    \end{subfigure}
    ~
    \begin{subfigure}[b]{0.3\textwidth}
        \includegraphics[width=\textwidth]{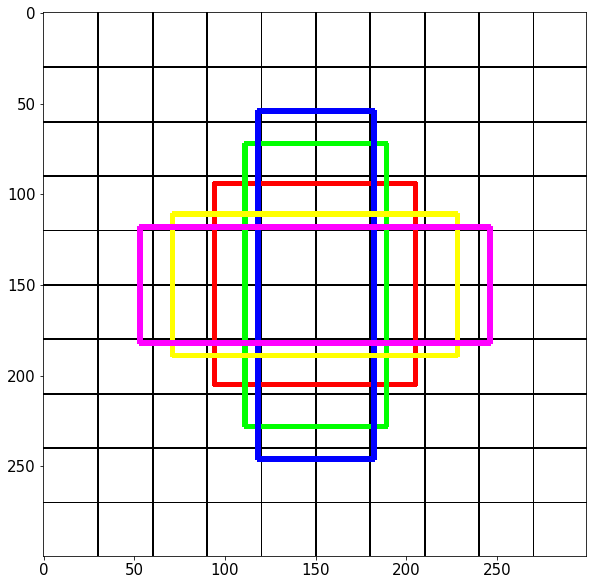}
        \caption{Prior anchor boxes for $10 \times 10$ feature map at conv conv 8\_2}
        \label{ssd_anchor_plot_default_2}
    \end{subfigure}
    \bigskip
    ~
    \begin{subfigure}[b]{0.3\textwidth}
        \includegraphics[width=\textwidth]{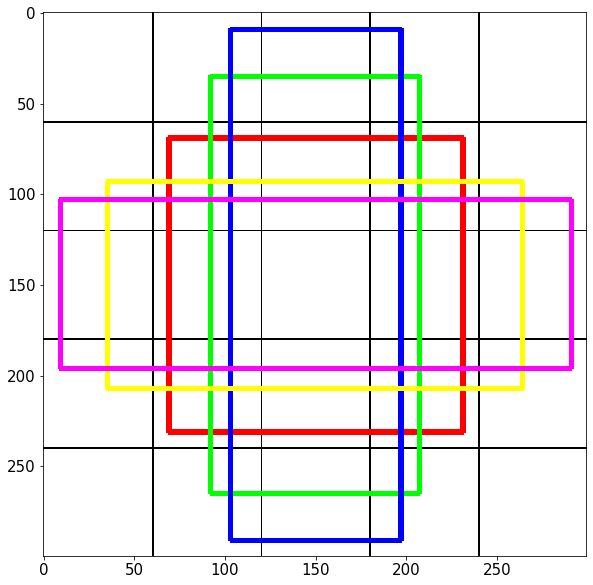}
        \caption{Prior anchor boxes for $5 \times 5$ feature map at  conv 9\_2}
        \label{ssd_anchor_plot_default_3}
    \end{subfigure}
    ~
    \begin{subfigure}[b]{0.3\textwidth}
        \includegraphics[width=\textwidth]{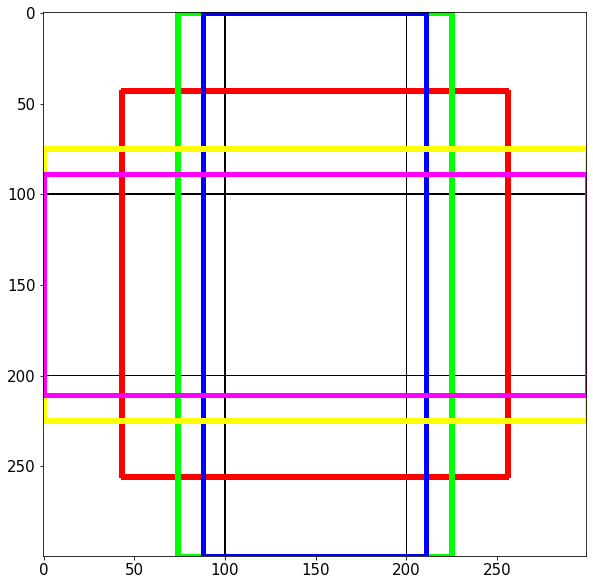}
        \caption{Prior anchor boxes for $3 \times 3$ feature map at conv 10\_2}
        \label{ssd_anchor_plot_default_4}
    \end{subfigure}
    ~
    \begin{subfigure}[b]{0.3\textwidth}
        \includegraphics[width=\textwidth]{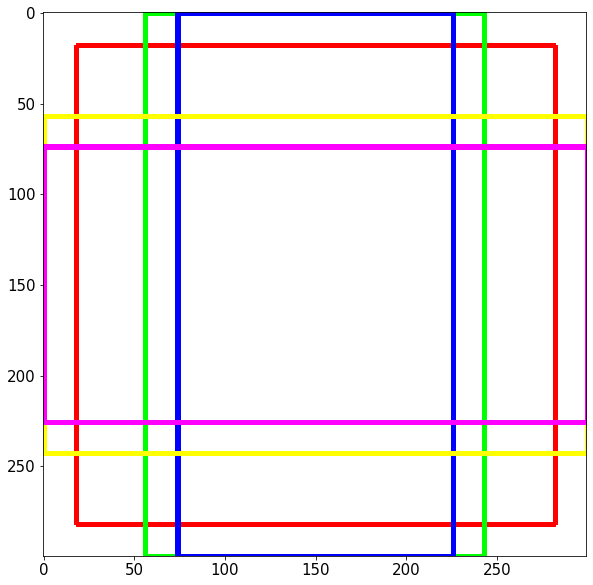}
        \caption{Prior anchor boxes for $1 \times 1$ feature map at conv 11\_2}
        \label{ssd_anchor_plot_default_5}
    \end{subfigure}
    \caption[Prior anchor boxes used in SSD]{Prior anchor boxes used in the original implementation of SSD on PASCAL VOC 2007}
    \label{ssd_anchors_overall}
\end{figure}

\begin{figure}[!htb]
    \centering
    \begin{subfigure}[b]{0.30\textwidth}
        \centering{BOGP}
    \end{subfigure}
    ~
    \begin{subfigure}[b]{0.30\textwidth}
        \centering{SMAC}
    \end{subfigure}
    ~
    \begin{subfigure}[b]{0.30\textwidth}
        \centering{CMA-ES}
    \end{subfigure}
    ~
    \begin{subfigure}[b]{0.30\textwidth}
        \includegraphics[width=\textwidth]{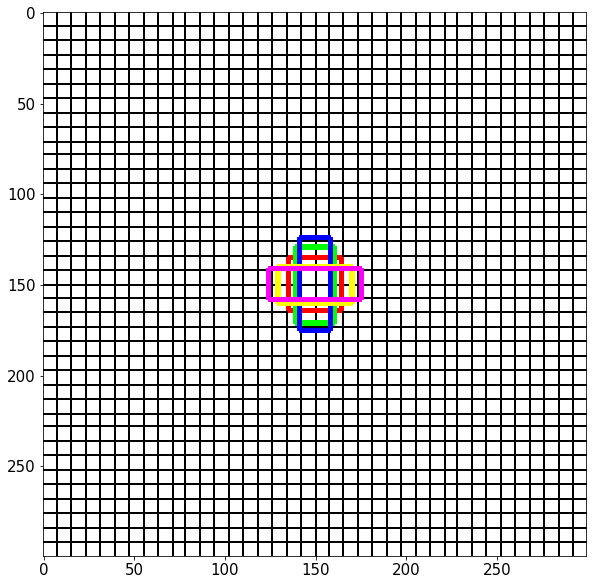}
        \label{ssd_anchor_plot_bogp_0}
    \end{subfigure}
    ~
    \begin{subfigure}[b]{0.30\textwidth}
        \includegraphics[width=\textwidth]{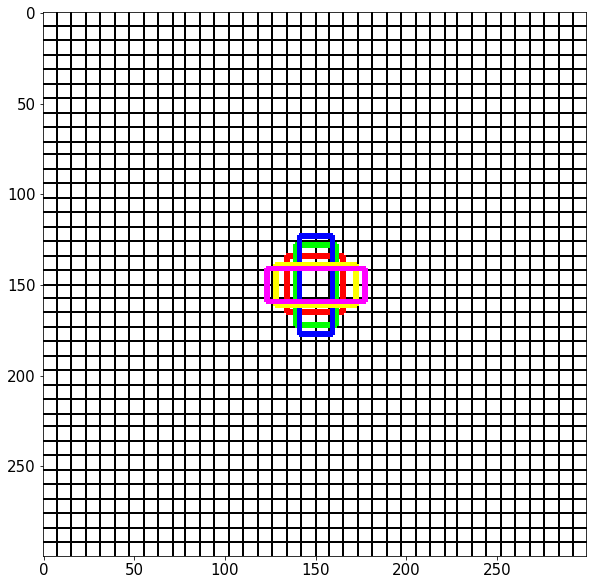}
        \label{ssd_anchor_plot_smac_0}
    \end{subfigure}
    ~
    \begin{subfigure}[b]{0.30\textwidth}
        \includegraphics[width=\textwidth]{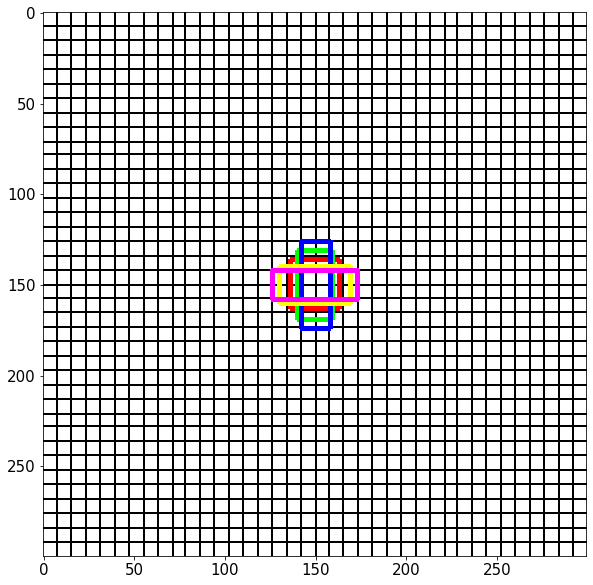}
        \label{ssd_anchor_plot_cmaes_0}
    \end{subfigure}
    \bigskip
    ~
    \begin{subfigure}[b]{0.30\textwidth}
        \includegraphics[width=\textwidth]{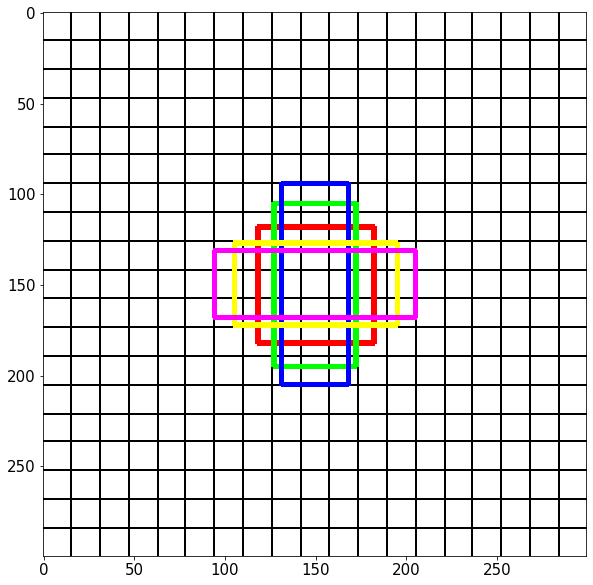}
        \label{ssd_anchor_plot_bogp_1}
    \end{subfigure}
    ~
    \begin{subfigure}[b]{0.30\textwidth}
        \includegraphics[width=\textwidth]{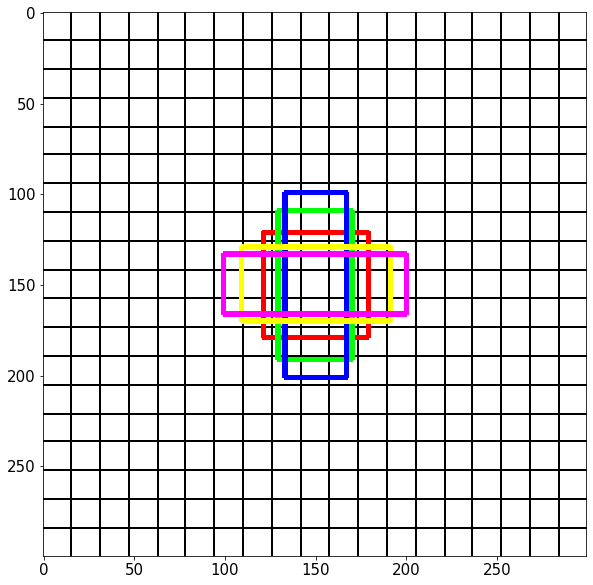}
        \label{ssd_anchor_plot_smac_1}
    \end{subfigure}
    ~
    \begin{subfigure}[b]{0.30\textwidth}
        \includegraphics[width=\textwidth]{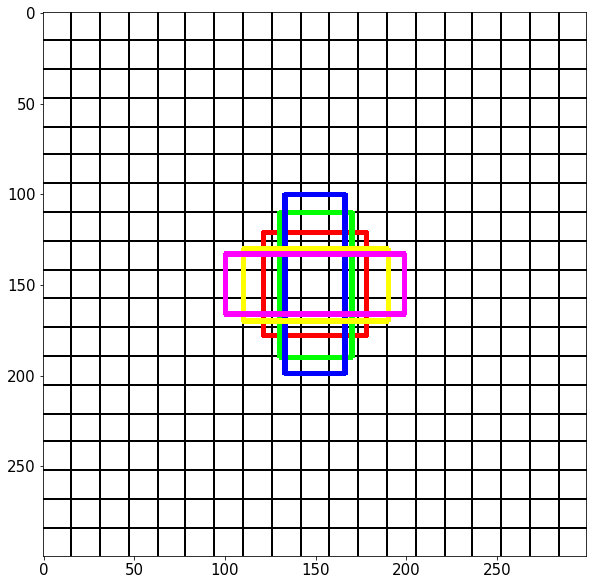}
        \label{ssd_anchor_plot_cmaes_1}
    \end{subfigure}
    \bigskip
    ~
    \begin{subfigure}[b]{0.30\textwidth}
        \includegraphics[width=\textwidth]{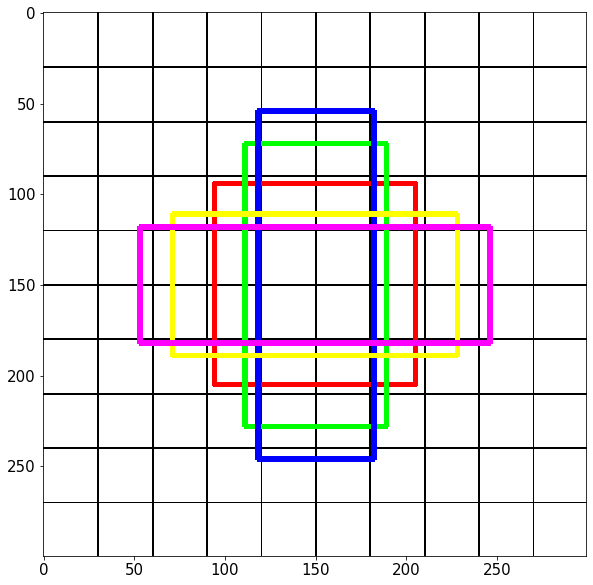}
        \label{ssd_anchor_plot_bogp_2}
    \end{subfigure}
    ~
    \begin{subfigure}[b]{0.30\textwidth}
        \includegraphics[width=\textwidth]{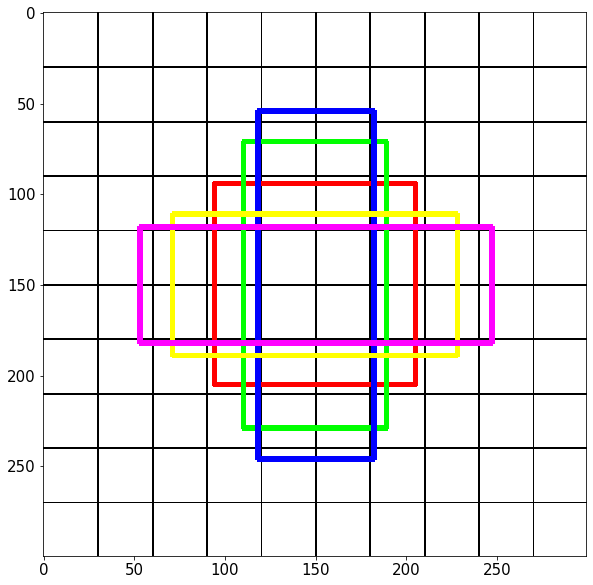}
        \label{ssd_anchor_plot_smac_2}
    \end{subfigure}
    ~
    \begin{subfigure}[b]{0.30\textwidth}
        \includegraphics[width=\textwidth]{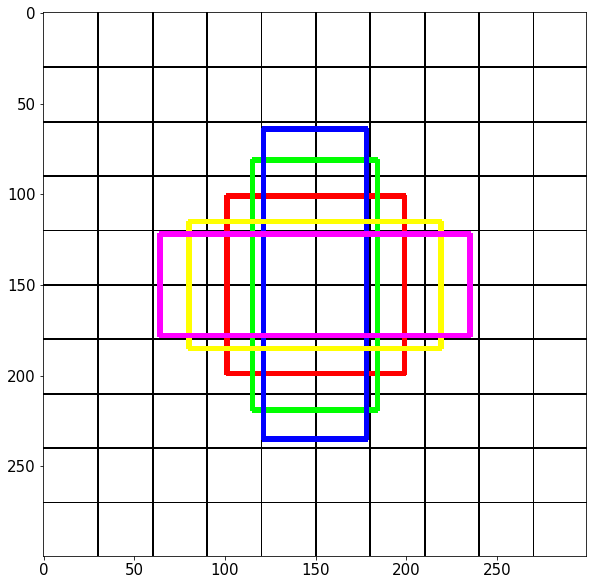}
        \label{ssd_anchor_plot_cmaes_2}
    \end{subfigure}
    \caption{Prior anchor boxes comparison of layers conv 4\_2, fc 7 and conv 8\_2 in SSD on PASCAL VOC 2007. First column: BOGP Second Column: SMAC, Third Column: CMA-ES}
    \label{ssd_anchors_overall_1}
\end{figure}

\begin{figure}[!htb]
    \centering
    \begin{subfigure}[b]{0.30\textwidth}
        \centering{BOGP}
    \end{subfigure}
    ~
    \begin{subfigure}[b]{0.30\textwidth}
        \centering{SMAC}
    \end{subfigure}
    ~
    \begin{subfigure}[b]{0.30\textwidth}
        \centering{CMA-ES}
    \end{subfigure}
    ~
    \begin{subfigure}[b]{0.30\textwidth}
        \includegraphics[width=\textwidth]{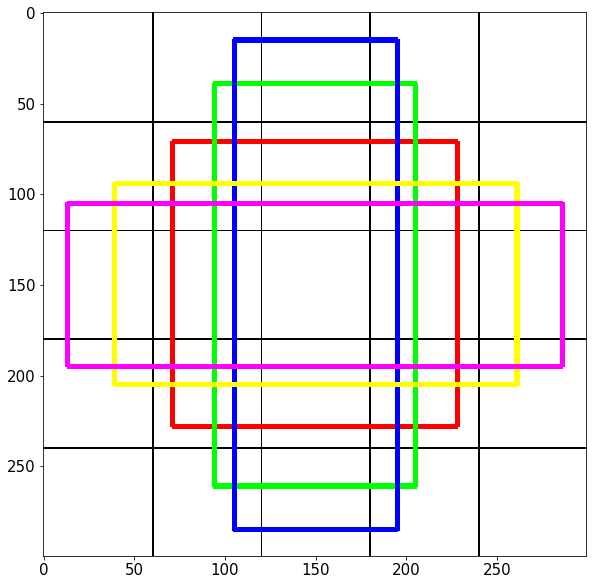}
        \label{ssd_anchor_plot_bogp_3}
    \end{subfigure}
    ~
    \begin{subfigure}[b]{0.30\textwidth}
        \includegraphics[width=\textwidth]{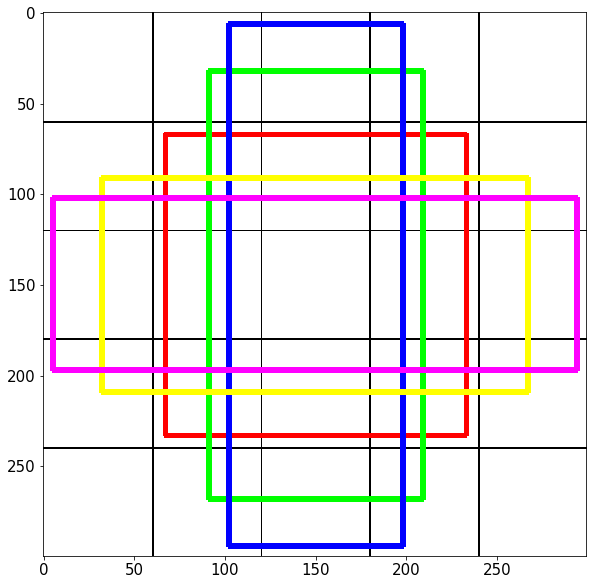}
        \label{ssd_anchor_plot_smac_3}
    \end{subfigure}
    ~
    \begin{subfigure}[b]{0.30\textwidth}
        \includegraphics[width=\textwidth]{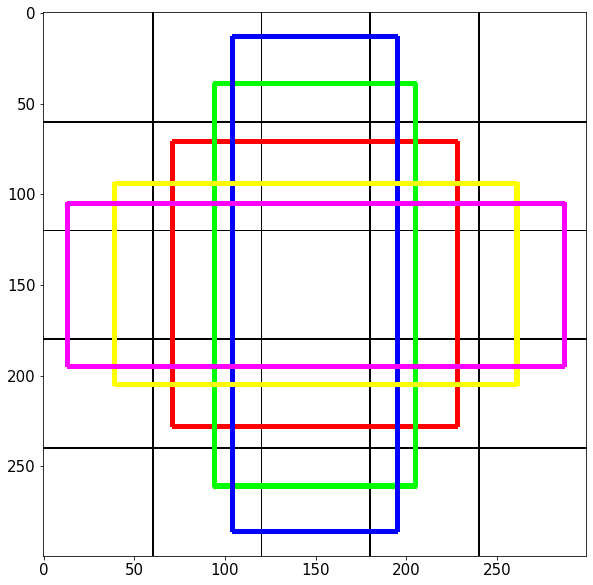}
        \label{ssd_anchor_plot_cmaes_3}
    \end{subfigure}
    \bigskip
    ~
    \begin{subfigure}[b]{0.30\textwidth}
        \includegraphics[width=\textwidth]{images/ssd_anchor_plot_default_4.png}
        \label{ssd_anchor_plot_default_4}
    \end{subfigure}
    ~
    \begin{subfigure}[b]{0.30\textwidth}
        \includegraphics[width=\textwidth]{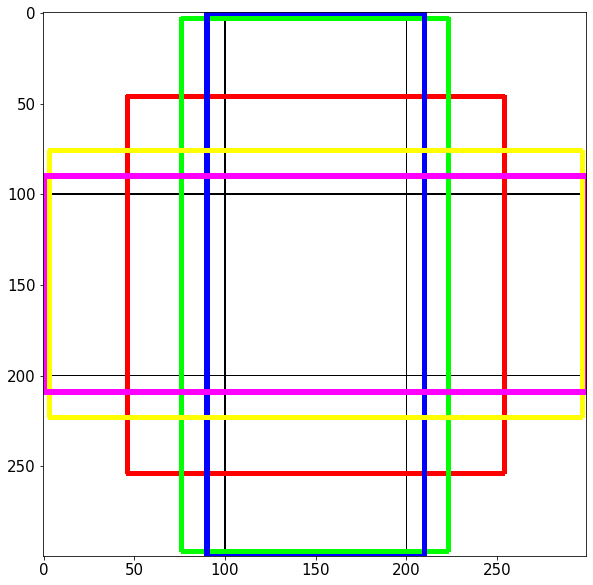}
        \label{ssd_anchor_plot_smac_4}
    \end{subfigure}
    ~
    \begin{subfigure}[b]{0.30\textwidth}
        \includegraphics[width=\textwidth]{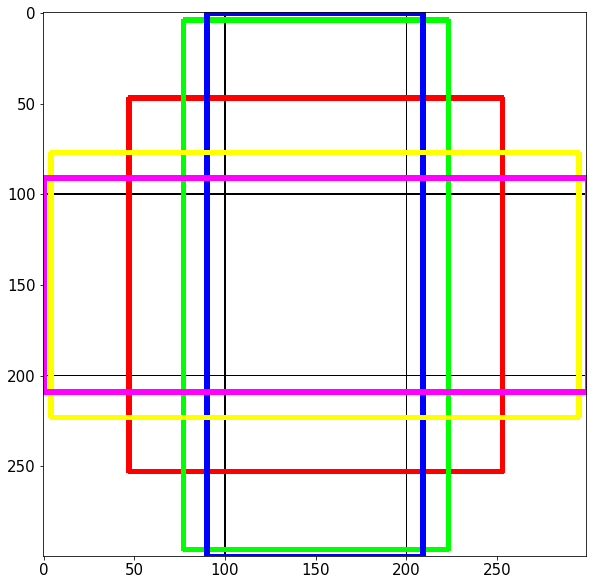}
        \label{ssd_anchor_plot_cmaes_4}
    \end{subfigure}
    \bigskip
    ~
    \begin{subfigure}[b]{0.30\textwidth}
        \includegraphics[width=\textwidth]{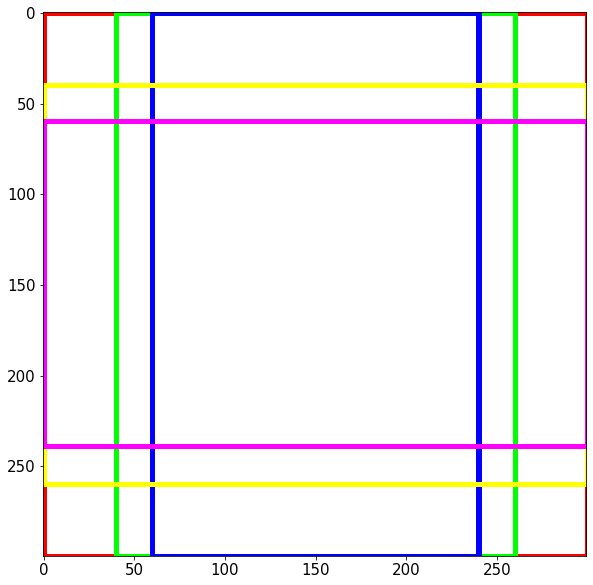}
        \label{ssd_anchor_plot_bogp_5}
    \end{subfigure}
    ~
    \begin{subfigure}[b]{0.30\textwidth}
        \includegraphics[width=\textwidth]{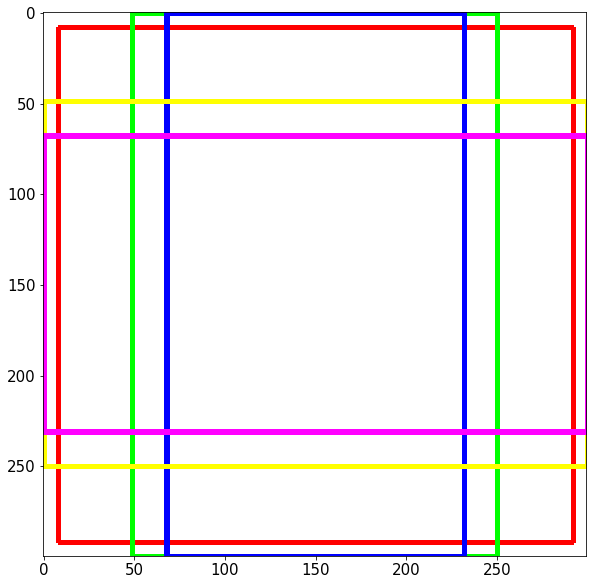}
        \label{ssd_anchor_plot_smac_5}
    \end{subfigure}
    ~
    \begin{subfigure}[b]{0.30\textwidth}
        \includegraphics[width=\textwidth]{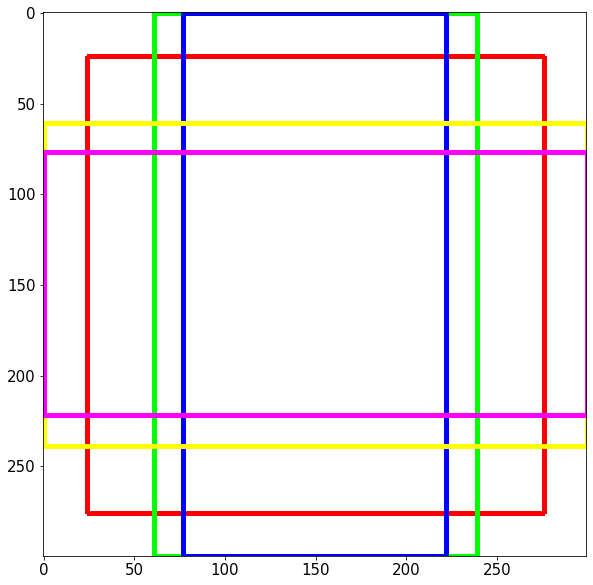}
        \label{ssd_anchor_plot_cmaes_5}
    \end{subfigure}
    \caption[Prior anchor boxes comparison deeper layers ]{Prior anchor boxes comparison of layers conv 9\_2, conv 10\_2 and conv 11\_2 in SSD on PASCAL VOC 2007. First column: BOGP Second Column: SMAC, Third Column: CMA-ES}
    \label{ssd_anchors_overall_2}
\end{figure}

Figures \ref{ssd_anchors_overall}, \ref{ssd_anchors_overall_1}, and \ref{ssd_anchors_overall_2} show a comparison of the learned anchor boxes with the original anchor boxes in SSD. The changes are very subtle, giving evidence that the original anchor boxes might be good enough for most use cases, but still there are large changes in the biggest scales (Figure \ref{ssd_anchors_overall_2}).

\FloatBarrier
\section{Additional Results on the Marine Debris Dataset}

An important motivation behind this work is to apply hyper-parameter optimization to real use cases that train object detectors in novel datasets. Considering this, we applied CMA-ES for tuning SSD anchor scales on a new marine debris dataset that we captured.

This dataset was created with the purpose of detecting and capturing debris with a marine surface vehicle. It is used to train object detectors, with preference for a detector with high mAP that runs in a resource-constrained computer inside the surface vehicle. The object classes required for this task are not well represented in most object detection datasets, specially for marine scenes, so we decided to build our own dataset. This dataset is not public yet, it will be released in a future publication that we are preparing.

\begin{figure}[!htb]
    \centering
    \includegraphics[width=0.8\linewidth]{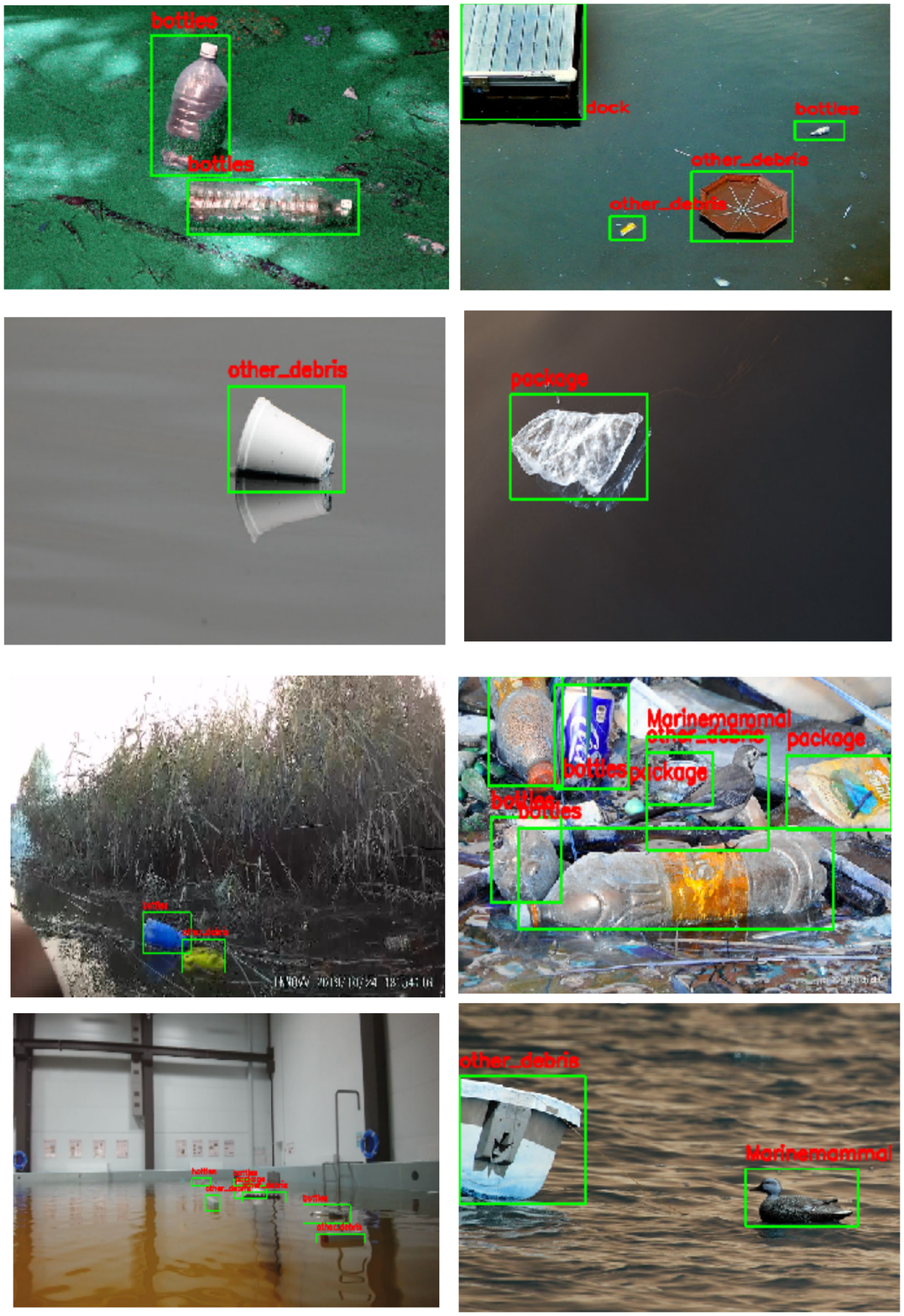}
    \captionsetup{justification=justified}
    \caption{Sample Images from the Marine Debris Dataset}
    \label{marine_sample_1}
\end{figure}

\begin{figure}[!htb]
    \centering
    \includegraphics[width=0.8\linewidth]{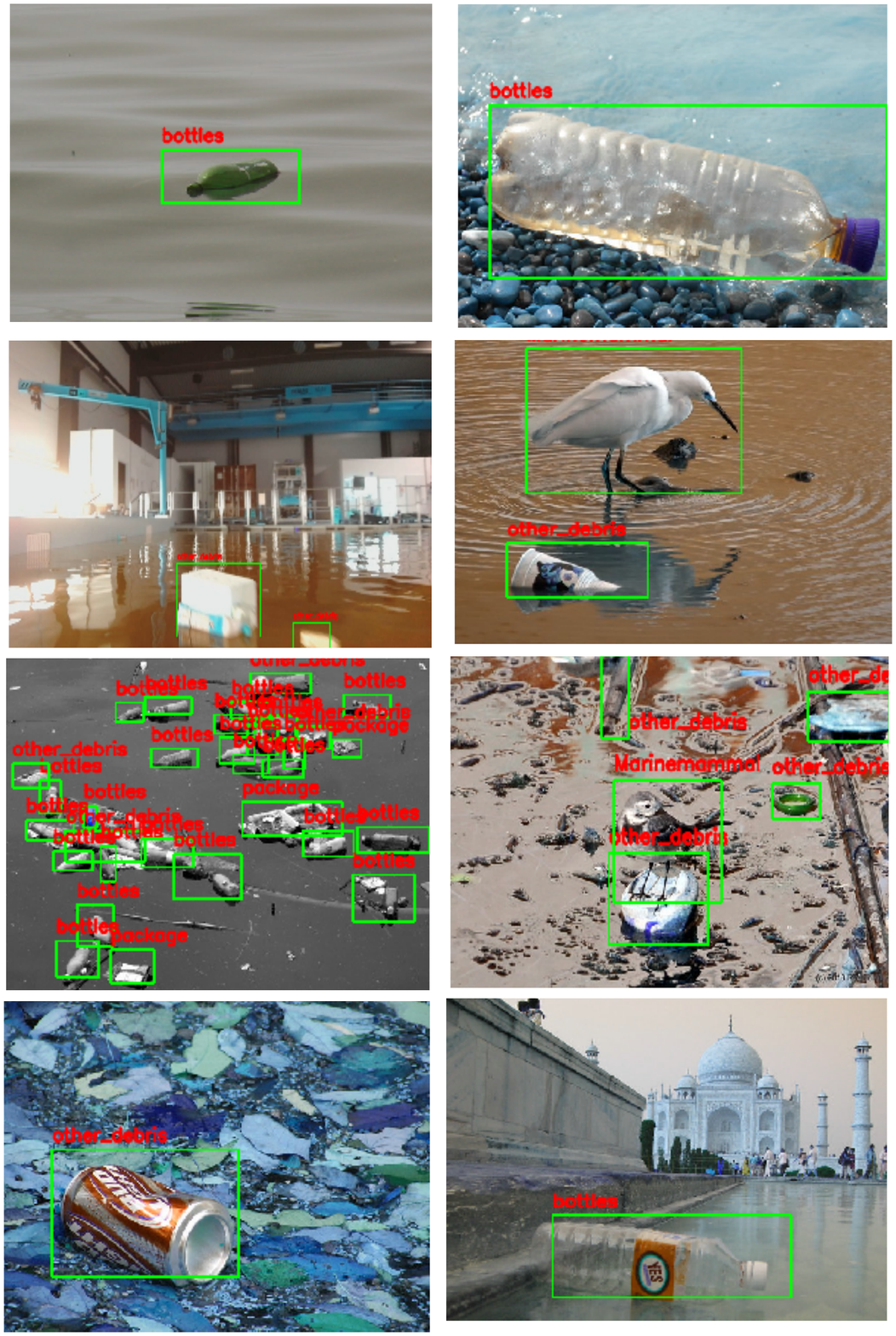}
    \captionsetup{justification=justified}
    \caption{Sample Images from the Marine Debris Dataset}
    \label{marine_sample_2}
\end{figure}

The images were collected and extracted from various online resources like TACO, the Open Images Dataset and Flickr Creative Commons Images, and some captured using a surface vehicle in a water tank. Annotation labels from TACO and Open Images are re-used and modified to suit our use case. We manually annotated other images coming from Flickr. Figure \ref{marine_sample_1} and \ref{marine_sample_2} shows some of the images from the Marine Debris Dataset.

The dataset consists of 2849 images for training, and 1079 images for evaluation. We annotated the following classes with bounding boxes:

\begin{itemize}
    \item Marine Vehicles: Ship, Boat, One person vehicle
    \item Humans: Swimmer
    \item Marine Structures: Pier, Dock, Bridge, Miscellaneous Structure
    \item Marine Animals: Marine Mammal
    \item Debris: Bottle, Package, Miscellaneous Debris, Wooden Debris, Debris Patch
\end{itemize}  

\begin{table}[!b]
    \centering
    \begin{center}
        \begin{tabular}{lllllll}
            \toprule
            \textbf{Dataset} & \textbf{\begin{tabular}[c]{@{}l@{}}Step\\ control\\ $\sigma$\end{tabular}} & \textbf{\begin{tabular}[c]{@{}l@{}}Population \\ Size\\ $\lambda$\end{tabular}} & \textbf{\begin{tabular}[c]{@{}l@{}}Maximum\\  Evaluations\end{tabular}} & \textbf{\begin{tabular}[c]{@{}l@{}}Number of \\ Genes\end{tabular}} & \textbf{\begin{tabular}[c]{@{}l@{}}Number of\\ Generations\end{tabular}} \\
            \midrule
            \begin{tabular}[c]{@{}l@{}}
                Marine \\ Debris \\ Dataset\end{tabular} & 0.3 & 8 & 112 & 7 & 14 \\
            \bottomrule
        \end{tabular}
    \end{center}
    \caption[CMA-ES initial parameter settings]{CMA-ES initial parameter settings on marine debris dataset}
    \label{tab:cmaes_initial_marine}
\end{table}

\clearpage

\subsection{Optimization of SSD Hyper-parameters using CMA-ES on the Marine Debris Dataset}

In this experiment, the prior anchor scales of SSD is optimized using CMA-ES. We use the same initial parameters and hyper-parameter space as the SSD experiments on PASCAL VOC. Table \ref{tab:cmaes_initial_marine} shows the CMA-ES parameters used in this experiment.

The hyper-parameter configurations generated in each generation is evaluated parallelly by training SSD on eight different Nvidia TitanXP GPUs. The model is initially trained with $10^{-3}$ learning rate for 20000 steps. Then, the training is continued with $10^{-4}$ and $10^{-5}$ learning rate for another 5000 steps, respectively. The batch size per training step is set to 32.

\begin{figure}[!tb]
    \centering
    \includegraphics[scale=0.20]{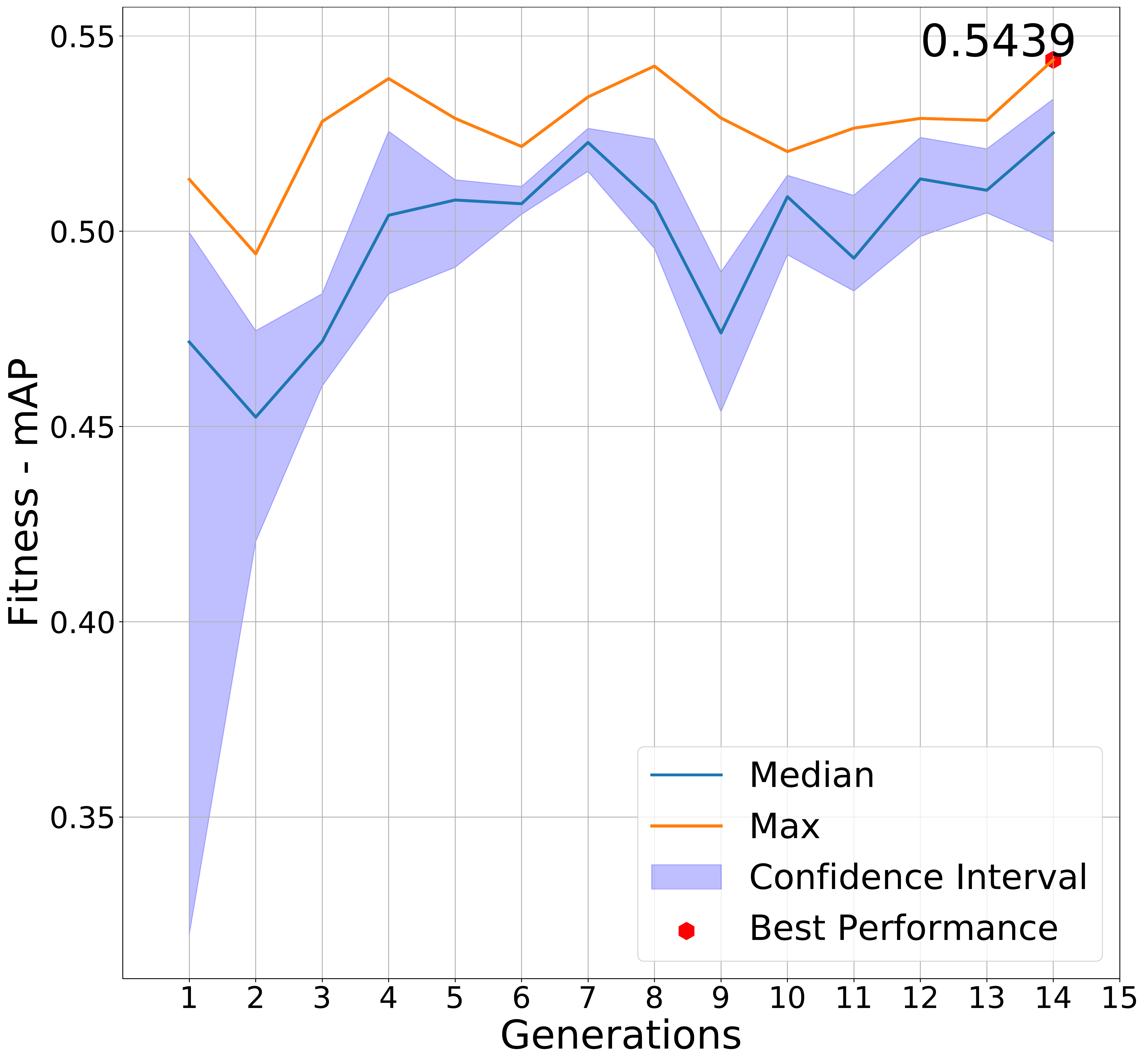}
    \caption[Optimization of SSD hyper-parameters (marine debris dataset)]{Optimization of SSD Hyper-parameters using CMA-ES on marine debris dataset (15 generations). The blue line indicates the median of mAP scores, while the orange line indicates the maximum of mAP scores of hyper-parameters evaluated in that particular generation. The increase in median over generations shows the evolution of distribution towards the better performing region. The red point shows the best hyper-parameter's mAP score.}
    \label{cmaes_results_ssd_marine}
\end{figure}

\begin{table}[!tb]
    \begin{center}
        \resizebox{0.85\textwidth}{!}{%
            \begin{tabular}{lllllllll}
                \toprule
                \textbf{Method} &  \multicolumn{7}{c}{\textbf{Anchor Scales}} & \textbf{mAP} \\ \cmidrule{2-8} 
                &   \multicolumn{1}{c}{\textbf{\begin{tabular}[c]{@{}l@{}}Scale \\ 0\end{tabular}}} & \multicolumn{1}{l}{\textbf{\begin{tabular}[c]{@{}l@{}}Scale \\ 1\end{tabular}}} & \multicolumn{1}{l}{\textbf{\begin{tabular}[c]{@{}l@{}}Scale \\ 2\end{tabular}}} & \multicolumn{1}{l}{\textbf{\begin{tabular}[c]{@{}l@{}}Scale \\ 3\end{tabular}}} & \multicolumn{1}{l}{\textbf{\begin{tabular}[c]{@{}l@{}}Scale \\ 4\end{tabular}}} & \multicolumn{1}{l}{\textbf{\begin{tabular}[c]{@{}l@{}}Scale \\ 5\end{tabular}}} & \multicolumn{1}{l}{\textbf{\begin{tabular}[c]{@{}l@{}}Scale\\  6\end{tabular}}} &
                \multicolumn{1}{l}{\textbf{\begin{tabular}[c]{@{}l@{}}\end{tabular}}} \\
                \midrule
                \textbf{Default} & 0.1 &  0.2&  0.37&  0.54&  0.71&  0.88& 1.05 & 51.47 \\
                \textbf{CMA-ES} & 0.04811& 0.2257& 0.3536&  0.5765& 0.7667&  0.8600& 1.03711  & \textbf{54.39} \\ 
                \bottomrule
            \end{tabular}
        }
    \end{center}
    
    \caption{Performance comparison of optimized SSD scales by different CMA-ES on marine debris datase. This demonstrates that scales tuned by CMA-ES achieves better results than the default configurations.}
    \label{tab:ssd_results_sum_marine}
\end{table}

\begin{figure}[!htb]
    \begin{center}
        \begin{subfigure}[b]{0.40\textwidth}
            \centering{Default}
        \end{subfigure}
        \begin{subfigure}[b]{0.40\textwidth}
            \centering{CMA-ES}
        \end{subfigure}
        
        \begin{subfigure}[b]{0.40\textwidth}
            \includegraphics[width=\textwidth]{images/ssd_anchor_plot_default_0.png}
            \label{ssd_anchor_plot_default_0}
        \end{subfigure}
        \begin{subfigure}[b]{0.40\textwidth}
            \includegraphics[width=\textwidth]{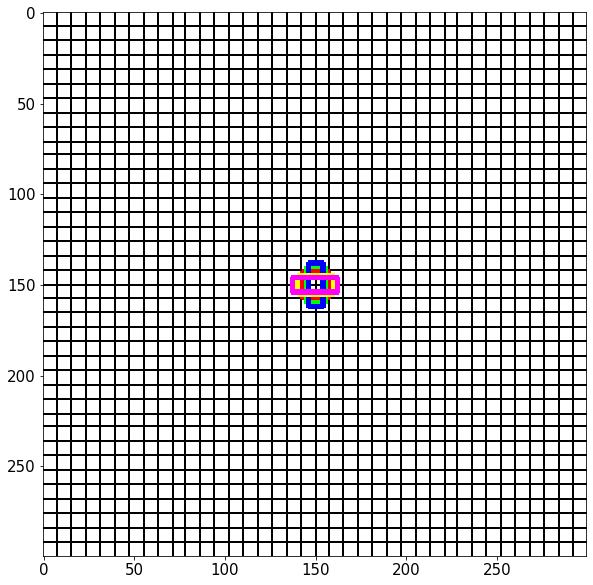}
            \label{ssd_anchor_plot_cmaes_marine_0}
        \end{subfigure}
        
        \begin{subfigure}[b]{0.40\textwidth}
            \includegraphics[width=\textwidth]{images/ssd_anchor_plot_default_1.png}
            \label{ssd_anchor_plot_default_1}
        \end{subfigure}
        \begin{subfigure}[b]{0.40\textwidth}
            \includegraphics[width=\textwidth]{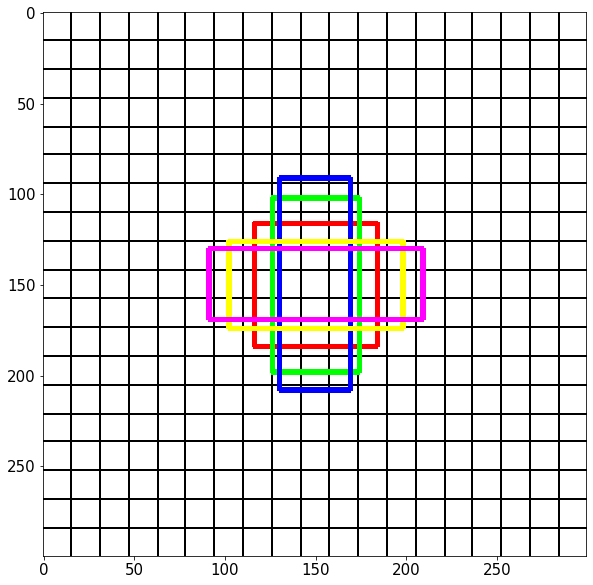}
            \label{ssd_anchor_plot_cmaes_marine_1}
        \end{subfigure}
        
        \begin{subfigure}[b]{0.40\textwidth}
            \includegraphics[width=\textwidth]{images/ssd_anchor_plot_default_2.png}
            \label{ssd_anchor_plot_default_2}
        \end{subfigure}
        \begin{subfigure}[b]{0.40\textwidth}
            \includegraphics[width=\textwidth]{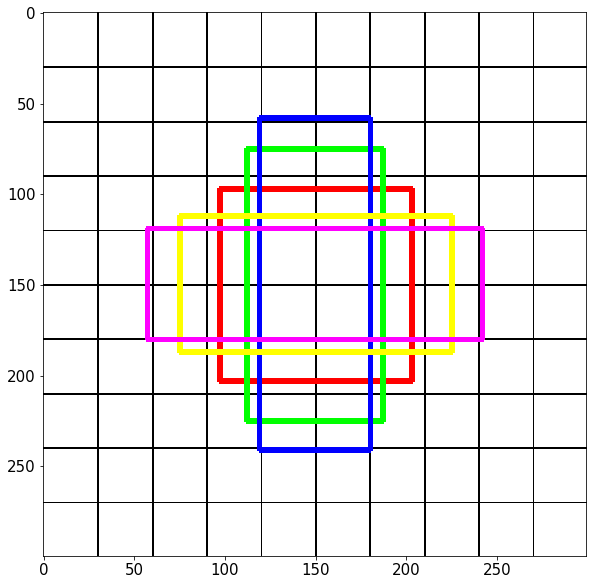}
            \label{ssd_anchor_plot_cmaes_marine_2}
        \end{subfigure}
    \end{center}
    \caption[Prior anchor boxes comparison - marine debris dataset]{Prior anchor boxes comparison in layers conv 4\_2, conv 7 and conv 8\_2 of SSD on the Marine Debris Dataset. First column: Default Implementation, Second Column: CMA-ES}
    \label{ssd_anchors_comp_debris_1}
\end{figure}

Figure \ref{cmaes_results_ssd_marine} shows the optimization of SSD anchor scales over generations. The best anchor scales found using CMA-ES achieves an mAP of 54.39\%, which is 2.92\% greater than the default scales. It is evident from Figure \ref{ssd_anchors_comp_debris_1} that the anchor scale 0 tuned by CMA-ES is smaller than the default anchor scale 0. Also there is a significant change in the anchor scale 4.

Moreover, the per-class average precision shown in Table \ref{comp_ap_marine} confirms the tuning of anchor scales has improved the performance of all objects except marine vehicle class. 

\begin{table}[!tb]
    \begin{center}
        \resizebox{0.75\textwidth}{!}{%
            \begin{tabular}{lllllll}
                \toprule
                \textbf{Method} & \textbf{\begin{tabular}[c]{@{}l@{}}Marine \\ structure \\ (AP)\end{tabular}} & \textbf{\begin{tabular}[c]{@{}l@{}}Marine \\ mammal\\ (AP)\end{tabular}} & \textbf{\begin{tabular}[c]{@{}l@{}}Debris\\ (AP)\end{tabular}} & \textbf{\begin{tabular}[c]{@{}l@{}}Marine \\ vehicle\\ (AP)\end{tabular}} & \textbf{\begin{tabular}[c]{@{}l@{}}Swimmer\\ (AP)\end{tabular}} & \textbf{mAP} \\
                \midrule
                \textbf{Default} & 0.55 & 0.62 & 0.49 & 0.48 & 0.44 & 51.47 \\
                \textbf{CMA-ES} & 0.58 & 0.65 & 0.51 & 0.47 & 0.49 & 54.39 \\
                \bottomrule
            \end{tabular}%
        }
    \end{center}
    \caption[Comparison of average precision of objects - Default vs CMA-ES]{Comparison of average precision of objects on marine debris dataset - Default vs CMA-ES}
    \label{comp_ap_marine}
\end{table}

\textbf{Regression analysis}. Similarly to previous experiments we perform regression analysis to find the important anchor scales. Table \ref{reg_ssd_marine} shows that anchor scales zero and six have more impact on mAP when compared to other anchor scales.

\begin{table}[!tb]
    \centering
    \begin{center}
        \begin{tabular}{lllllllll}
            \toprule
            \textbf{Method} & $\mathbf{R^2}$ & \multicolumn{7}{c}{\textbf{Coefficients}} \\ \cmidrule{3-9} 
            &  & \multicolumn{1}{c}{\textbf{\begin{tabular}[c]{@{}l@{}}Scale \\ 0\end{tabular}}} & \multicolumn{1}{l}{\textbf{\begin{tabular}[c]{@{}l@{}}Scale \\ 1\end{tabular}}} & \multicolumn{1}{l}{\textbf{\begin{tabular}[c]{@{}l@{}}Scale \\ 2\end{tabular}}} & \multicolumn{1}{l}{\textbf{\begin{tabular}[c]{@{}l@{}}Scale \\ 3\end{tabular}}} & \multicolumn{1}{l}{\textbf{\begin{tabular}[c]{@{}l@{}}Scale \\ 4\end{tabular}}} & \multicolumn{1}{l}{\textbf{\begin{tabular}[c]{@{}l@{}}Scale \\ 5\end{tabular}}} & \multicolumn{1}{l}{\textbf{\begin{tabular}[c]{@{}l@{}}Scale\\  6\end{tabular}}} \\
            \midrule
            CMA-ES 	& 0.4 & 0.37 & 0.03 & 0.18 & 0.02 & 0.03 & 0.14 & 0.41\\
            \bottomrule
        \end{tabular}
    \end{center}
    \caption{Regression analysis of SSD hyper-parameters on the Marine Debris Dataset}
    \label{reg_ssd_marine}
\end{table}	

\begin{figure}[!htb]
    \begin{center}
        \begin{subfigure}[b]{0.40\textwidth}
            \centering{Default}
        \end{subfigure}
        \begin{subfigure}[b]{0.40\textwidth}
            \centering{CMA-ES}
        \end{subfigure}
        
        \begin{subfigure}[b]{0.40\textwidth}
            \includegraphics[width=\textwidth]{images/ssd_anchor_plot_default_3.png}
            \label{ssd_anchor_plot_default_3}
        \end{subfigure}
        \begin{subfigure}[b]{0.40\textwidth}
            \includegraphics[width=\textwidth]{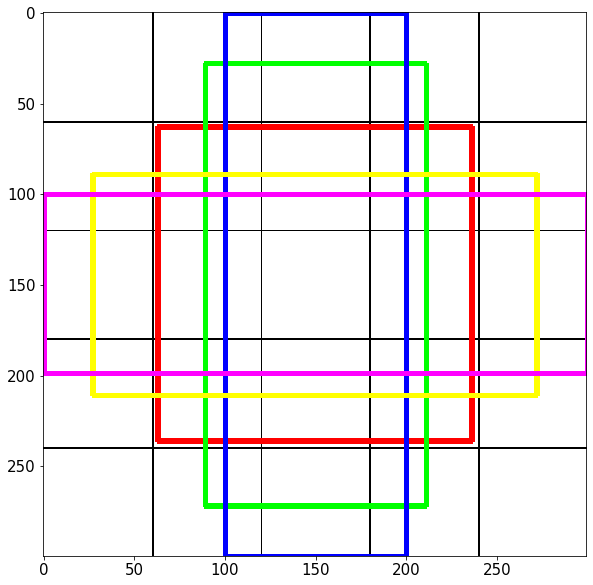}
            \label{ssd_anchor_plot_cmaes_marine_3}
        \end{subfigure}
        
        \begin{subfigure}[b]{0.40\textwidth}
            \includegraphics[width=\textwidth]{images/ssd_anchor_plot_default_4.png}
            \label{ssd_anchor_plot_default_4}
        \end{subfigure}
        \begin{subfigure}[b]{0.40\textwidth}
            \includegraphics[width=\textwidth]{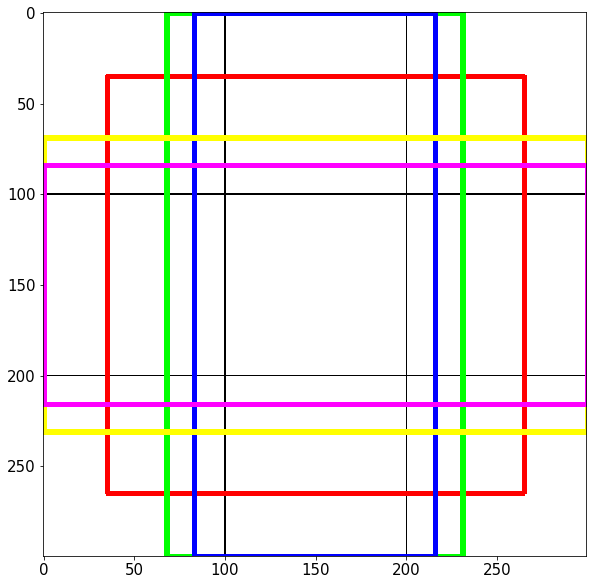}
            \label{ssd_anchor_plot_cmaes_marine_4}
        \end{subfigure}

        \begin{subfigure}[b]{0.40\textwidth}
            \includegraphics[width=\textwidth]{images/ssd_anchor_plot_default_5.png}
            \label{ssd_anchor_plot_default_5}
        \end{subfigure}
        \begin{subfigure}[b]{0.40\textwidth}
            \includegraphics[width=\textwidth]{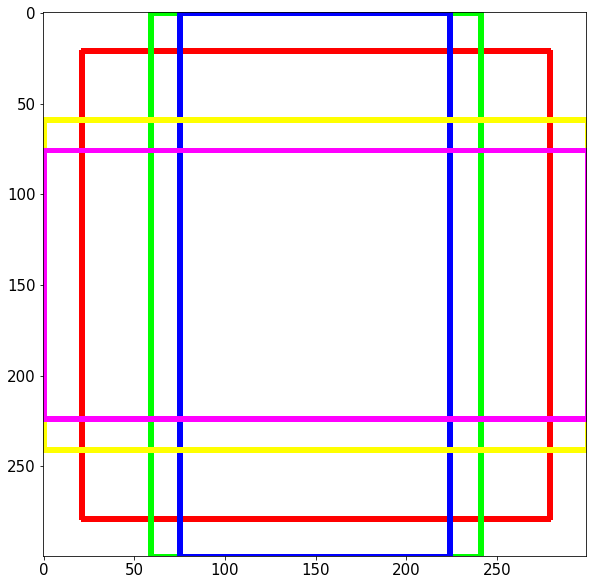}
            \label{ssd_anchor_plot_cmaes_marine_5}
        \end{subfigure}
    \end{center}
    \caption[Prior anchor boxes comparison - marine debris dataset]{Prior anchor boxes comparison in layers conv9\_2, conv10\_2, and conv 11\_2 of SSD on Marine Debris Dataset. First column: Default Implementation, Second Column: CMA-ES}
    \label{ssd_anchors_comp_debris_2}
\end{figure}

\textbf{Scales visualization}. In Figures \ref{ssd_anchors_comp_debris_1} and \ref{ssd_anchors_comp_debris_2} we present a comparison the scales that have been learned in this dataset using CMA-ES with the original SSD scales. The differences are small but there are relevant changes, which connect to our regression analysis results in Table \ref{reg_ssd_marine}.

\end{document}